\documentclass[10pt,twocolumn,letterpaper]{article}

\usepackage{cvpr}
\usepackage{color,soul}
\usepackage{times}
\usepackage{epsfig}
\usepackage{graphicx}
\usepackage{amsmath}
\usepackage{amssymb}
\usepackage{enumitem}
\usepackage{caption}
\usepackage{subcaption}
\usepackage{booktabs}
\usepackage[export]{adjustbox}
\usepackage{makecell}
\usepackage{mathtools}
\usepackage{authblk}

\usepackage[pagebackref=true,breaklinks=true,letterpaper=true,colorlinks,bookmarks=false]{hyperref}

\cvprfinalcopy %

\newcommand{\za}{\ensuremath{z^a}}

\makeatletter
\renewcommand\paragraph{\@startsection{paragraph}{4}{\z@}%
  {.75ex \@plus1ex \@minus .2ex}%
  {-1em}  %
  {\reset@font\normalsize\bfseries}
  }
\renewcommand\AB@affilsepx{, \protect\Affilfont}
\makeatother

\def\cvprPaperID{4943}

\ifcvprfinal\fi
\begin{document}
\setlength{\belowcaptionskip}{-10pt}

\title{Neural Rerendering in the Wild}

\author[1]{Moustafa Meshry\thanks{Work performed during an internship at Google.}}
\author[2]{Dan B Goldman}
\author[2]{Sameh Khamis}
\author[2]{Hugues Hoppe}
\author[2]{\mbox{Rohit Pandey}}
\author[2]{\mbox{Noah Snavely}}
\author[2]{Ricardo Martin-Brualla}
\affil[1]{University of Maryland}\affil[2]{Google Inc.}

\renewcommand\Authands{, }

\maketitle

\begin{abstract}

We explore total scene capture --- recording, modeling, and rerendering a scene under varying appearance such as season and time of day.
Starting from internet photos of a tourist landmark, we apply traditional 3D reconstruction to register the photos and approximate the scene as a point cloud.
For each photo, we render the scene points into a deep framebuffer,
and train a neural network to learn the mapping of these initial renderings to the actual photos.
This rerendering network also takes as input a latent appearance vector and a semantic mask indicating the location of transient objects like pedestrians.
The model is evaluated on several datasets of publicly available images spanning a broad range of illumination conditions.
We create short videos demonstrating realistic manipulation of the image viewpoint, appearance, and semantic labeling.
We also compare results with prior work on scene reconstruction from internet photos.

\end{abstract}

\section{Introduction}

Imagine spending a day sightseeing in Rome in a fully realistic interactive experience without ever stepping on a plane.
One could visit the Pantheon in the morning, enjoy the sunset overlooking the Colosseum, and fight through the crowds to admire the Trevi Fountain at night time.
Realizing this goal involves capturing the complete appearance space of a scene, \ie, recording a scene under all possible lighting conditions and transient states in which the scene might be observed---be it crowded, rainy, snowy, sunrise, spotlit, etc.---and then being able to summon up any viewpoint of the scene under any such condition. 
We call this ambitious vision \emph{total scene capture}.
It is extremely challenging due to the sheer diversity of appearance---scenes can look dramatically different under night illumination, during special events, or in extreme weather.

In this paper, we focus on capturing tourist landmarks around the world using publicly available community photos as the sole input, \ie, photos \emph{in the wild}.
Recent advances in 3D reconstruction can generate impressive 3D models from such photo collections~\cite{agarwal2009building,shan2013visual,snavely2006photo}, but the renderings produced from the resulting point clouds or meshes lack the realism and diversity of real-world images.
Alternatively, one could use webcam footage to record a scene at regular intervals but without viewpoint diversity,
or use specialized acquisition (\eg, Google Street View, aerial, or satellite images) to snapshot the environment over a short time window but without appearance diversity.
In contrast, community photos offer an abundant (but challenging) sampling of appearances of a scene over many years.

\begin{figure}
    \centering
    \captionsetup[subfigure]{aboveskip=1pt,belowskip=1pt}
    \begin{subfigure}[b]{0.49\linewidth}
        \includegraphics[width=\linewidth]{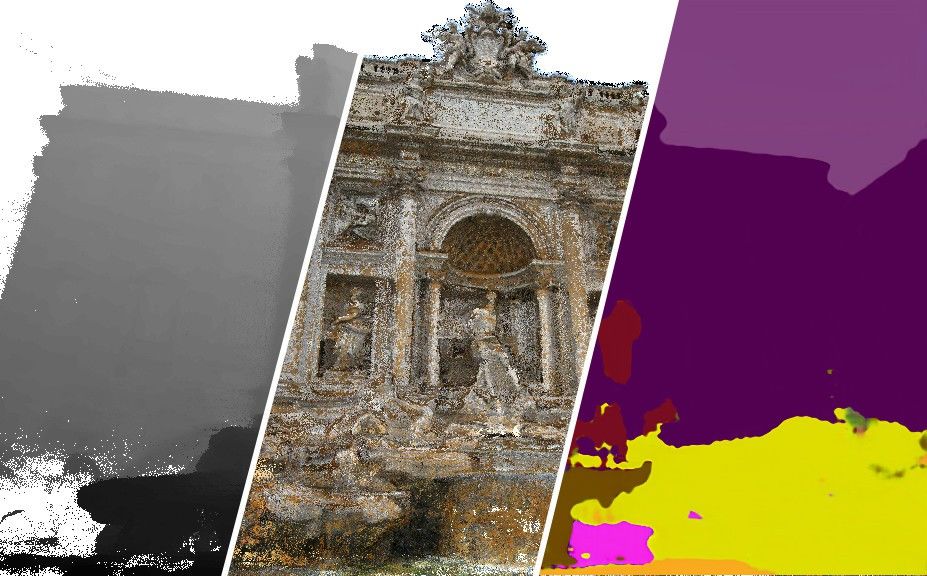}%
        \vspace*{.03in}  %
        \caption{Input deep buffer}
    \end{subfigure} 
    \begin{subfigure}[b]{0.49\linewidth}
        \includegraphics[width=\linewidth]{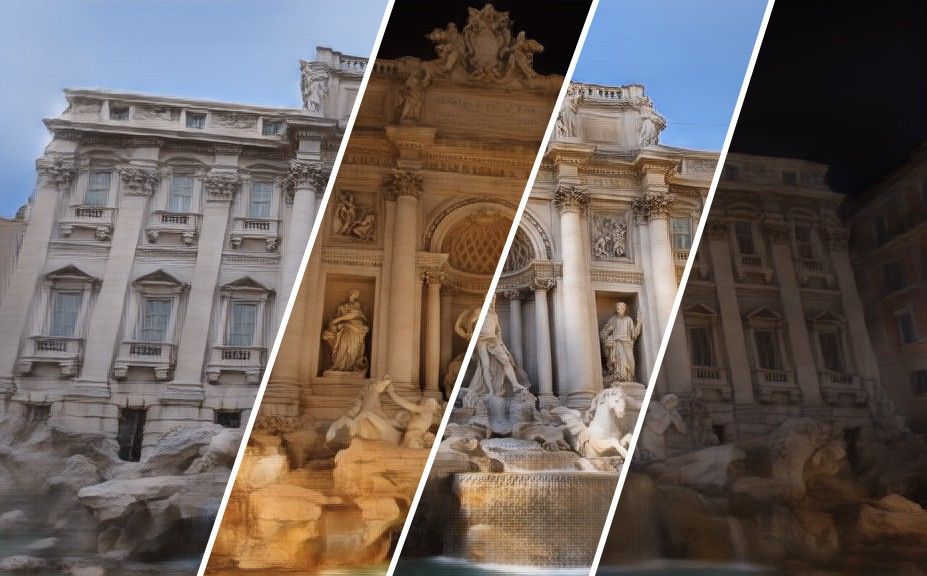}%
        \vspace*{.03in}  %
        \caption{Output renderings}
    \end{subfigure}%
    \vspace*{.03in}  %
    \caption{Our neural rerendering technique uses a large-scale internet photo collection to reconstruct a proxy 3D model and trains a neural rerendering network that takes as input a deferred-shading deep buffer (consisting of depth, color and semantic labeling) generated from the proxy 3D model (left), and outputs realistic renderings of the scene under multiple appearances (right).}
    \label{fig:teaser}
    \vspace*{.07in}  %
\end{figure}

Our approach to total scene capture has two main components:
(1)~creating a factored representation of the input images, which separates viewpoint, appearance conditions, and transient objects such as pedestrians, and
(2)~rendering realistic images from this factored representation.
Unlike recent approaches that extract implicit disentangled representations of viewpoint and content~\cite{park2017transformation,rhodin2018unsupervised,tatarchenko2016multi},
we employ state-of-the-art reconstruction methods to create an explicit intermediate 3D representation, in the form of a dense but noisy point cloud, and use this 3D representation as a ``scaffolding'' to predict images.

An explicit 3D representation lets us cast the rendering problem as a multimodal image translation~\cite{huang2018multimodal, lee2018diverse,zhu2017toward}.
The input is a deferred-shading framebuffer~\cite{saito1990comprehensible} in which each rendered pixel stores albedo, depth, and other attributes,
and the outputs are realistic views under different appearances.
We train the model by generating paired datasets, using the recovered viewpoint parameters of each input image to render a deep buffer of the scene from the same view, \ie, with pixelwise alignment.
Our model effectively learns to take an approximate initial scene rendering and rerender a realistic image.
This is similar to recent neural rerendering frameworks~\cite{kim2018deepvideoportraits,martinbrualla2018lookingood,thies2018ignor} but using uncontrolled internet images rather than carefully captured footage.

We explore a novel strategy to train the multimodal image translation model.
Rather than jointly estimating an embedding space for the appearance together with the rendering network~\cite{huang2018multimodal,lee2018diverse,zhu2017toward}, our system performs staged training of both. 
First, an appearance encoding network is pretrained using a proxy style-based loss~\cite{gatys2016image}, an efficient way to capture the style of an image. 
Then, the rerendering network is trained with fixed appearance embeddings from the pretrained encoder.
Finally, both the appearance encoding and rerendering networks are jointly finetuned.
This simple yet effective strategy lets us train simpler networks on large datasets.
We demonstrate experimentally how a model trained in this fashion better captures scene appearance.

Our system is a first step towards addressing total scene capture and focuses primarily on the static parts of scenes. 
Transient objects (\eg, pedestrians and cars) are handled by conditioning the rerendering network on the expected semantic labeling of the output image, so that the network can learn to ignore these objects rather than trying to hallucinate their locations.
This semantic labeling is also effective at discarding 
small or thin scene features (\eg, lampposts) whose geometry cannot be robustly reconstructed, yet are easily identified using image segmentation methods. 
Conditioning our network on a semantic mask also enables the rendering of scenes free of people if desired.
Code will be available at {\small\url{https://bit.ly/2UzYlWj}}.

In summary, our contributions include:

\begin{itemize}[noitemsep,nolistsep]
\item A first step towards total scene capture, \ie, recording and rerendering a scene under any appearance from in-the-wild photo collections.
\item A factorization of input images into viewpoint, appearance, and semantic labeling, conditioned on an approximate 3D scene proxy, from which we can \emph{rerender} realistic views under varying appearance.
\item A more effective method to learn the appearance latent space by pretraining the appearance embedding network using a proxy loss.
\item Compelling results including view and appearance interpolation on five large datasets, and direct comparisons to previous methods~\cite{shan2013visual}.
\end{itemize}

\section{Related work}

\paragraph{Scene reconstruction}

Traditional methods for scene reconstruction first generate a sparse reconstruction using large-scale structure-from-motion~\cite{agarwal2009building}, then perform Multi-View Stereo (MVS)~\cite{furukawa2010accurate, schoenberger2016mvs} or variational optimization~\cite{hiep2009towards} to reconstruct dense scene models. 
However, most such techniques assume a single appearance, or else simply recover an average appearance of the scene.  We build upon these techniques, using dense point clouds recovered from MVS as proxy geometry for neural rerendering.

In image-based rendering~\cite{debevec1996modeling,gortler1996lumigraph}, input images are used to generate new viewpoints by warping input pixels into the outputs using proxy geometry.
Recently, Hedman~\etal~\cite{hedman2018deepblending} introduce a neural network to compute blending weights for view-dependent texture mapping that reduces artifacts in poorly reconstructed regions.
However, image-based rendering generally assumes the captured scene has static appearance, so it is not well-suited to our problem setup in which the appearance varies across images.

Neural scene rendering~\cite{eslami2018neural} applies deep neural networks to learn a latent scene representation that allows generation of novel views, but is limited to simple synthetic geometry.

\paragraph{Appearance modeling}

A given scene can have dramatically different appearances at different times of day, in different weather conditions, and can also change over the years. Garg~\etal~\cite{garg2009dimensionality} observe that for a given viewpoint, the dimensionality of scene appearance as captured by internet photos is relatively low, with the exception of outliers like transient objects. One can recover illumination models for a photo collection by estimating albedo using cloudy images~\cite{shan2013visual}, retrieving the sun's location through timestamps and geolocation~\cite{hauagge_bmvc2014_outdoor}, estimating coherent albedos across the collection~\cite{laffont2012coherent}, or assuming a fixed viewpoint~\cite{sunkavalli2007factored}. However, these methods assume simple lighting models that do not apply to nighttime scene appearance. Radenovic~\etal~\cite{radenovic2016from} recover independent day and night reconstructions, but do not enable smooth appearance interpolations between the two.

Laffont~\etal~\cite{laffont2014transient} assign transient attributes like ``fall'' or ``sunny'' to each image, and learn a database of patches that allows for editing such attributes. Other works require direct supervision from lighting models estimated using 360-degree images~\cite{HoldGeoffroy2017DeepOI}, or ground truth object geometry~\cite{wang2018joint}.
In contrast, we use a data-driven implicit representation of appearance that is learned from the input image distribution and does not require direct supervision.

\paragraph{Deep image synthesis}

The seminal work of pix2pix~\cite{isola2017image} trains a deep neural network to translate an image from one domain, such as a semantic labeling, into another domain, such as a realistic image, using paired training data. Image-to-image (I2I) translation has since been applied to many tasks~\cite{dong2017semantic,ledig2017photo,pathak2016context,wang2016generative,wang2018video,zhang2017age}.
Several works propose improvements to stabilize training and allow for high-quality image synthesis~\cite{karras2017progressive,wang2017high,wang2018video}.
Others extend the I2I framework to unpaired settings where images from two domains are not in correspondence~\cite{kim2017learning,liu2017unsupervised,zhu2017unpaired}, multimodal outputs where an input image can map to multiple images~\cite{wang2017high,zhu2017toward}, or unpaired datasets with multimodal outputs where an image in one domain is converted to another domain while preserving the content~\cite{almahairi2018augmented,huang2018multimodal,lee2018diverse}.

Image translation techniques can be used to rerender scenes in a more realistic domain, to enable facial expression synthesis~\cite{kim2018deepvideoportraits}, to fix artifacts in captured 3D performances~\cite{martinbrualla2018lookingood}, or to add viewpoint-dependent effects~\cite{thies2018ignor}. 
In our paper, we demonstrate an approach for training a neural rerendering framework \emph{in the wild}, \ie, with uncontrolled data instead of captures under constant lighting conditions. We cast this as a multimodal image synthesis problem, where a given viewpoint can be rendered under multiple appearances using a latent appearance vector, and with editable semantics by conditioning the output on the desired semantic labeling of the output.

\section{Total scene capture}

We define the problem of \emph{total scene capture} as creating a generative model for all images of a given scene.
We would like such a model to:
\begin{itemize}[noitemsep,nolistsep]
    \item[--] encode the 3D structure of the scene, enabling rendering from an arbitrary viewpoint,  
    \item[--] capture all possible appearances of the scene, \eg, all lighting and weather conditions, and allow rendering the scene under any of them, and
    \item[--] understand the location and appearance of transient objects in the scene, \eg, pedestrians and cars, and allow for reproducing or omitting them.
\end{itemize}
Although these goals are ambitious, we show that one can create such a generative model given sufficient images of a scene, such as those obtained for popular tourist landmarks. %

We first describe a neural rerendering framework that we adapt from previous work in controlled capture settings~\cite{martinbrualla2018lookingood} to the more challenging setting of unstructured photo collections (Section~\ref{sec:neural_rerendering_framework}). We extend this model to enable appearance capture and multimodal generation of renderings under different appearances (Section~\ref{sec:appearance_modeling}).  We further extend the model to handle transient objects in the training data by conditioning its inputs on a semantic labeling of the ground truth images (Section~\ref{sec:semantic_conditioning}).

\subsection{Neural rerendering framework}
\label{sec:neural_rerendering_framework}

We adapt recent neural rerendering frameworks~\mbox{\cite{kim2018deepvideoportraits,martinbrualla2018lookingood}} to work with unstructured photo collections. Given a large internet photo collection $\{I_i\}$ of a scene, we first generate a proxy 3D reconstruction using COLMAP~\cite{schonberger2016colmap,schoenberger2016sfm,schoenberger2016mvs}, 
which applies Structure-from-Motion (SfM) and Multi-View Stereo (MVS) to create a dense colored point cloud.

An alternative to a point cloud is to generate a textured mesh~\cite{kazhdan2006poisson,waechter2014let}. Although meshes generate more complete renderings, they tend to also contain pieces of misregistered floating geometry which can occlude large regions of the scene~\cite{shan2013visual}. As we show later, our neural rerendering framework can produce highly realistic images given only point-based renderings as input.

Given the proxy 3D reconstruction, we generate an aligned dataset of rendered images and real images by rendering the 3D point cloud from the viewpoint $v_i$ of each input image $I_i$, where $v_i$ consists of camera intrinsics and extrinsics recovered via SfM.
We generate a deferred-shading deep buffer $B_i$ for each image~\cite{saito1990comprehensible}, which may contain per-pixel albedo, normal, depth and any other derivative information. In our case, we only use albedo and depth and render the point cloud by using point splatting with a z-buffer with a radius of 1 pixel.%

\begin{figure}[t]
    \centering
    \includegraphics[width=\linewidth]{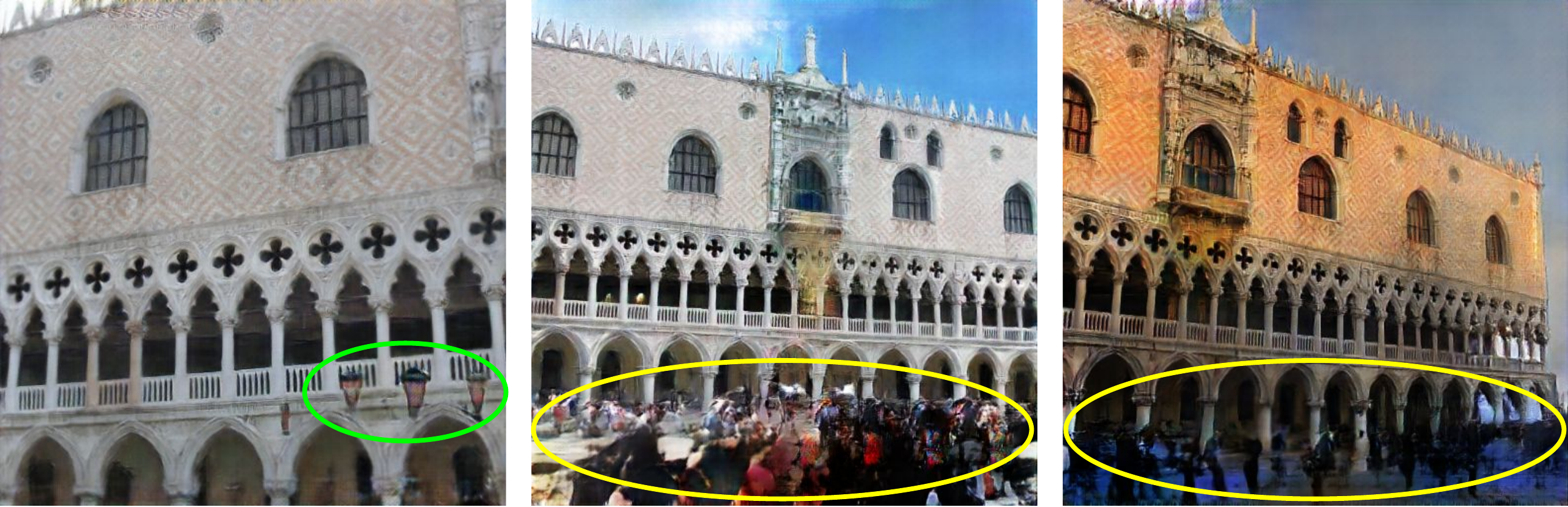}
    \caption{Output frames of a standard image translation network~\cite{isola2017image} trained for neural rerendering in a small dataset of 250 photos of San Marco.
    The network overfits the dataset and learns to hallucinate lampposts close to their approximate location in the scene (green), and virtual tourists (yellow), as well as memorizing a per-viewpoint appearance matching the specific input photos.}
    \label{fig:pix2pix_limitations}
\end{figure}

However, the image-to-image translation paradigm used in~\cite{kim2018deepvideoportraits,martinbrualla2018lookingood} is not appropriate for our use case, as it assumes a one-to-one mapping between inputs and outputs. 
A scene observed from a particular viewpoint can look very different depending on weather, lighting conditions, color balance, post processing filters, etc. 
In addition, a one-to-one mapping fails to explain transient objects in the scene, such as pedestrians or cars, whose location and individual appearance is impossible to predict from the static scene geometry alone.
Interestingly, if one trains a sufficiently large neural network on this simple task on a dataset, the network learns to (1) associate viewpoint with appearance via memorization and (2) hallucinate the location of transient objects, as shown in Figure~\ref{fig:pix2pix_limitations}.

\begin{figure}[t]
    \centering
    \includegraphics[width=\linewidth]{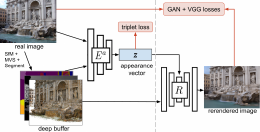}\\
    \caption{ An aligned dataset is created using Structure from Motion (SfM) and Multi-View Stereo (MVS). Our staged approach pre-trains the appearance encoder $E^a$ using a triplet loss (left). Then the rerenderer $R$ is trained using standard reconstruction and GAN losses (right), and finally fine-tuned together with $E^a$. \footnotesize{Photo Credits Rafael Jimenez (Creative Commons)}.}
    \label{fig:arch}
    \vspace*{.02in}
\end{figure}

\subsection{Appearance modeling}
\label{sec:appearance_modeling}
To capture the one-to-many relationship between input viewpoints (represented by their deep buffers $B_i$) and output images $I_i$ under different appearances, we cast the rerendering task as multimodal image translation~\cite{zhu2017toward}. 
In such a formulation, the goal is to learn a latent appearance vector $\za_i$ that captures variations in the output domain $I_i$ that cannot be inferred from the input domain $B_i$.
We compute the latent appearance vector as $\za_i = E^a(I_i, B_i)$ where $E^a$ is an appearance encoder that takes as input both the output image $I_i$ and the deep buffer $B_i$. 
We argue that having the appearance encoder $E^a$ observe the input $B_i$ allows it to  learn more complex appearance models by correlating the lighting in $I_i$ with scene geometry in $B_i$.
Finally, a rerendering network $R$ generates a scene rendering conditioned on both viewpoint $B_i$ and the latent appearance vector $\za$.
Figure~\ref{fig:arch} shows an overview of the overall process.

To train the appearance encoder $E^a$ and rendering network $R$, we first adopted elements from recent methods in multimodal synthesis~\cite{huang2018multimodal,lee2018diverse,zhu2017toward} to find a combination that is most effective in our scenario. 
However, this combination still has shortcomings as it is unable to model infrequent appearances well.
For instance, it does not reliably capture night appearances for scenes in our datasets.
We hypothesize that the appearance encoder (which is jointly trained with the rendering network) is not expressive enough to capture the large variability in the data.

To improve the model expressiveness, our approach is to stabilize the joint training of $R$ and $E^a$ by pretraining the appearance network $E^a$ independently on a proxy task.
We then employ a staged training approach in which the rendering network $R$ is first trained using fixed appearance embeddings, and finally we jointly fine-tune both networks. 
This staged training regime allows for a simpler model that captures more complex appearances.

We present our baseline approach, which adapts state-of-the-art multimodal synthesis techniques, and then our staged training strategy, which pretrains the appearance encoder on a proxy task.

\paragraph{Baseline}
Our baseline uses BicycleGAN~\cite{zhu2017toward} with two main adaptations.
First, our appearance encoder also takes as input the buffer~$B_i$, as described above.
Second, we add a cross-cycle consistency loss similar to~\cite{huang2018multimodal,lee2018diverse} to encourage appearance transfer across viewpoints.
Let $\za_1 = E^a(I_1, B_1)$ be the captured appearance of an input image $I_1$. 
We apply a reconstruction loss between image~$I_1$ and cross-cycle reconstruction~$\hat{I}_1 = R(B_1, \hat{z}^a_1)$, where $\hat{z}^a_1$ is computed through a cross-cycle with a second image $(I_2, B_2)$, i.e. $\hat{z}^a_1 = E^a(R(B_2, \za_1)), B_2)$.
We also apply a GAN loss on the intermediate appearance transfer output $R(B_2,\za_1)$ as in~\cite{huang2018multimodal,lee2018diverse}.

\begin{figure*}
    \centering
    \captionsetup[subfigure]{labelformat=empty,aboveskip=1pt,belowskip=0pt}

    \begin{subfigure}[b]{0.135\textwidth}
        \includegraphics[height=2.3cm]{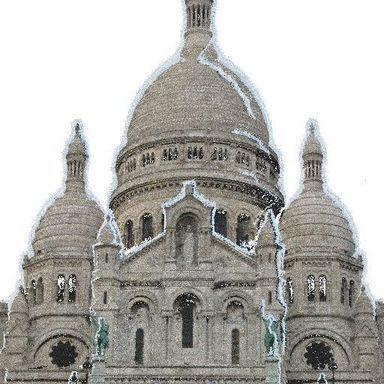}
    \end{subfigure}%
    \begin{subfigure}[b]{0.135\textwidth}
        \includegraphics[height=2.3cm]{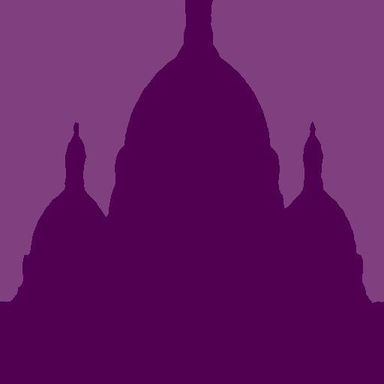}
    \end{subfigure}%
    \begin{subfigure}[b]{0.135\textwidth}
        \includegraphics[height=2.3cm]{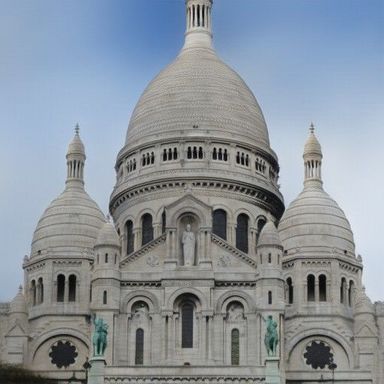}
    \end{subfigure}%
    \begin{subfigure}[b]{0.135\textwidth}
        \includegraphics[height=2.3cm]{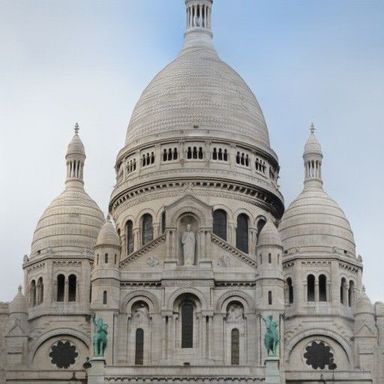}
    \end{subfigure}%
    \begin{subfigure}[b]{0.135\textwidth}
        \includegraphics[height=2.3cm]{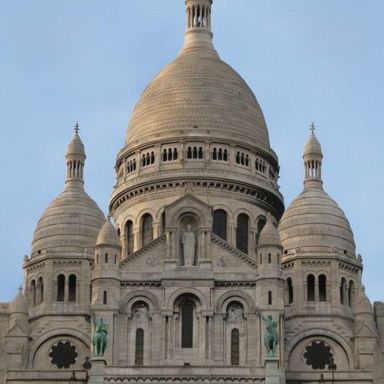}
    \end{subfigure}%
    \begin{subfigure}[b]{0.135\textwidth}
        \includegraphics[height=2.3cm]{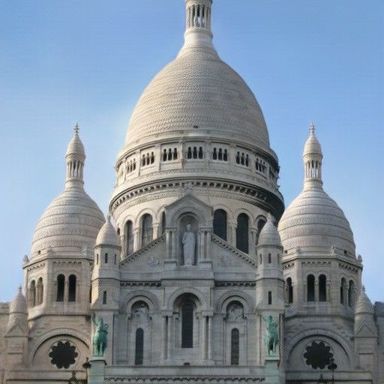}
    \end{subfigure}%
    \begin{subfigure}[b]{0.135\textwidth}
        \includegraphics[height=2.3cm]{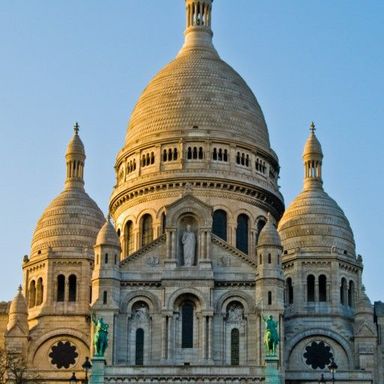}
    \end{subfigure}
    \begin{subfigure}[b]{0.135\textwidth}
        \includegraphics[height=2.3cm]{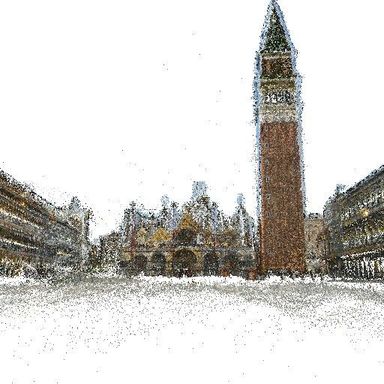}
    \end{subfigure}%
    \begin{subfigure}[b]{0.135\textwidth}
        \includegraphics[height=2.3cm]{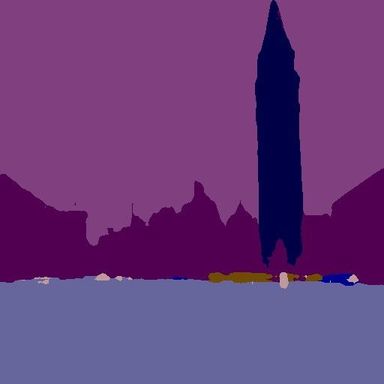}
    \end{subfigure}%
    \begin{subfigure}[b]{0.135\textwidth}
        \includegraphics[height=2.3cm]{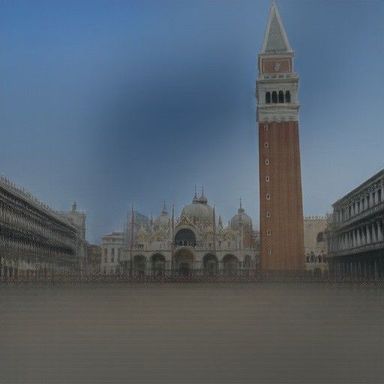}
    \end{subfigure}%
    \begin{subfigure}[b]{0.135\textwidth}
        \includegraphics[height=2.3cm]{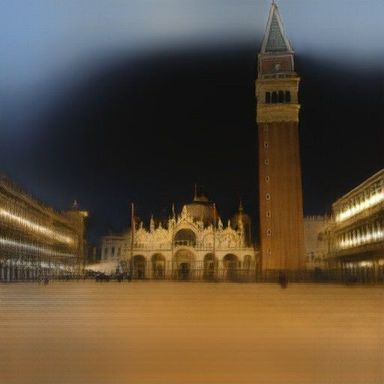}
    \end{subfigure}%
    \begin{subfigure}[b]{0.135\textwidth}
        \includegraphics[height=2.3cm]{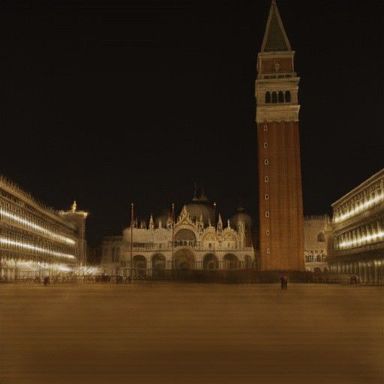}
    \end{subfigure}%
    \begin{subfigure}[b]{0.135\textwidth}
        \includegraphics[height=2.3cm]{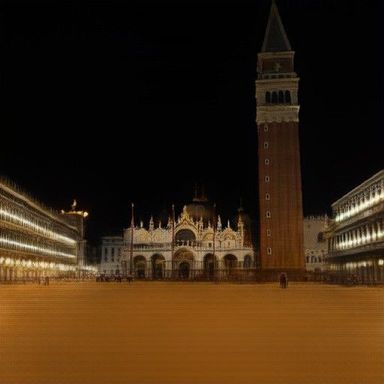}
    \end{subfigure}%
    \begin{subfigure}[b]{0.135\textwidth}
        \includegraphics[height=2.3cm]{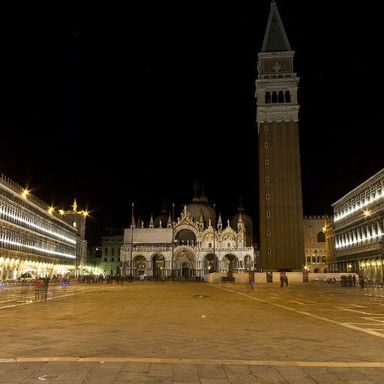}
    \end{subfigure}\\
    \begin{subfigure}[b]{0.135\textwidth}
        \includegraphics[height=2.3cm]{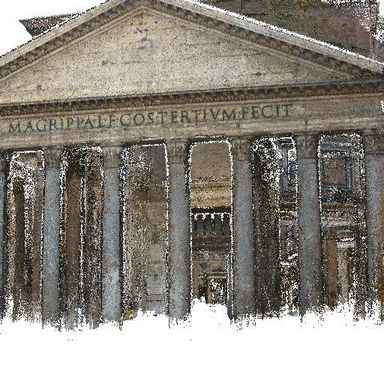}
        \caption{\hspace{-0.5em}Input}
    \end{subfigure}%
    \begin{subfigure}[b]{0.135\textwidth}
        \includegraphics[height=2.3cm]{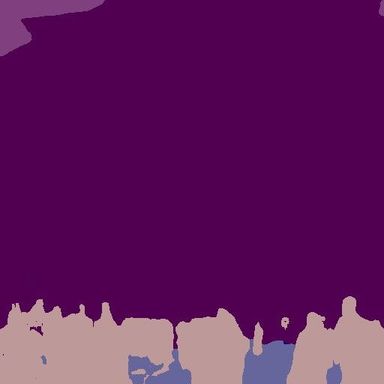}
        \caption{\hspace{-0.5em}Segmentation}
    \end{subfigure}%
    \begin{subfigure}[b]{0.135\textwidth}
        \includegraphics[height=2.3cm]{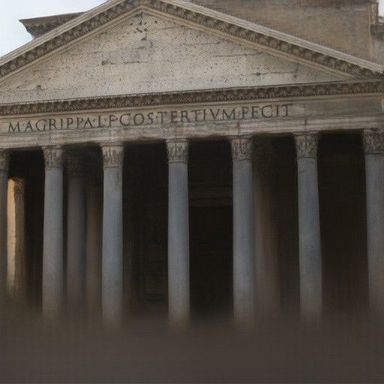}
        \caption{\hspace{-0.5em}I2I}
    \end{subfigure}%
    \begin{subfigure}[b]{0.135\textwidth}
        \includegraphics[height=2.3cm]{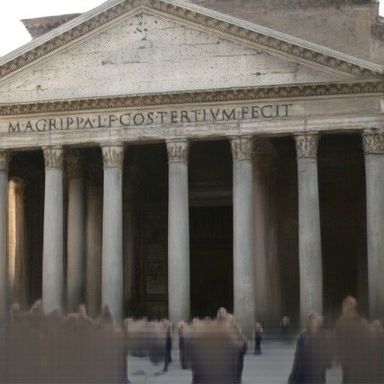}
        \caption{\hspace{-0.5em}+Sem}
    \end{subfigure}%
    \begin{subfigure}[b]{0.135\textwidth}
        \includegraphics[height=2.3cm]{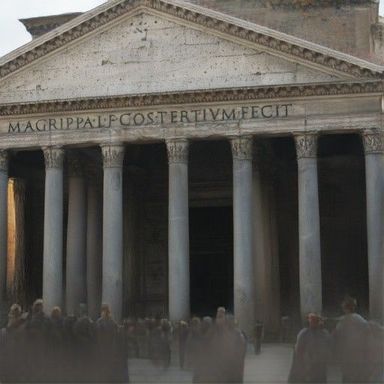}
        \caption{\hspace{-0.5em}+Sem+BaseApp}
    \end{subfigure}%
    \begin{subfigure}[b]{0.135\textwidth}
        \includegraphics[height=2.3cm]{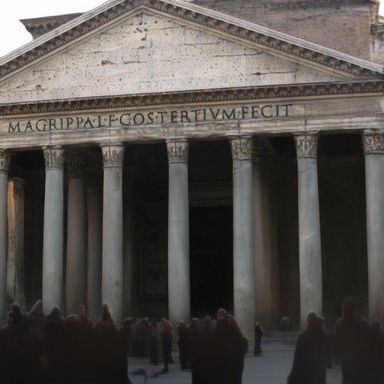}
        \caption{\hspace{-0.5em}+Sem+StagedApp}
    \end{subfigure}%
    \begin{subfigure}[b]{0.135\textwidth}
        \includegraphics[height=2.3cm]{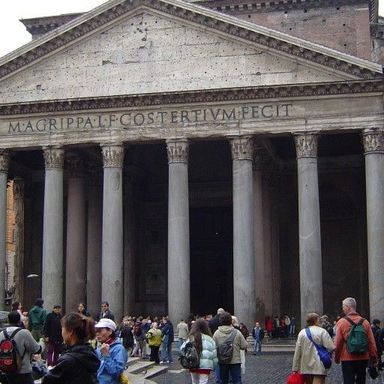}
        \caption{\hspace{-0.5em}Ground Truth}
    \end{subfigure}\\
    \caption{Example visual results of our ablative study in Table~\ref{tab:quantitative_results}.
    From left to right, input color render, segmentation mask from the corresponding ground truth images, result using an image-to-image baseline (I2I), with semantic conditioning (+Sem), and with semantic conditioning and a baseline appearance modeling based on~\cite{zhu2017toward} (+Sem+BaseApp), with semantic conditioning and staged appearance training (+Sem+StagedApp).
    \footnotesize{Photo Credits: Flickr users Gary Campbell-Hall, Steve Collis, and Tahbepet (Creative Commons).}}\label{fig:ablative_study}
\end{figure*}

\paragraph{Staged appearance training}

The key to our staged training approach is the appearance pretraining stage, where we pretrain the appearance encoder $E^a$ independently on a proxy task.
We then train the rendering network $R$ while fixing the weights of $E^a$, allowing $R$ to find the correlations between output images and the embedding produced by the proxy task.
Finally, we fine-tune both $E^a$ and $R$ jointly.

This staged approach simplifies and stabilizes the training of $R$, enabling training of a simpler network with fewer regularization terms.
In particular, we remove the cycle and cross-cycle consistency losses, the latent vector reconstruction loss, and the KL-divergence loss, leaving only a direct reconstruction loss and a GAN loss.
We show experimentally in Section~\ref{sec:evaluation} that this approach results in better appearance capture and rerenderings than the baseline model.

\paragraph{Appearance pretraining}
To pretrain the appearance encoder $E^a$, we choose a proxy task that optimizes an embedding of the input images into the appearance latent space using a suitable distance metric between input images.
This training encourages embeddings such that if two images are close under the distance metric, then their appearance embeddings should also be close in the appearance latent space. 
Ideally the distance metric we choose should ignore the content or viewpoint of $I_i$ and $B_i$, as our goal is to encode a latent space that is independent of viewpoint.
Experimentally we find that the style loss employed in neural style-transfer work~\cite{gatys2016image} has such a property; it largely ignores content and focuses on more abstract properties.

\setlength{\abovedisplayskip}{3pt}
\setlength{\belowdisplayskip}{3pt}

To train the embedding, we use a triplet loss, where for each image $I_i$, we find the set of $k$ closest and furthest neighbor images given by the style loss, from which we can sample a positive sample $I_{p}$ and negative sample $I_{n}$, respectively.
The loss is then:
\[
\mathcal{L}(I_i, I_{p}, I_{n}) = \sum_j \max\left(\|g_i^j\!-\!g_{p}^j\|^2 - \|g_i^j\!-\!g_{n}^j\|^2 + \alpha, 0\right)
\]
where $g^j_i$ is the Gram matrix of activations at the $j^{th}$ layer of a VGG network %
of image $I_i$, and $\alpha$ is a separation margin.

\subsection{Semantic conditioning}
\label{sec:semantic_conditioning}

To account for transient objects in the scene, we condition the rerendering network on a semantic labeling $S_i$ of image $I_i$ that depicts the location of transient objects such as pedestrians.
Specifically,
we concatenate the semantic labeling $S_i$ to the deep buffer $B_i$ wherever the deep buffer was previously used.
This discourages the network from encoding variations caused by the location of transient objects in the appearance vector, or associating such transient objects with specific viewpoints, as shown in Figure~\ref{fig:pix2pix_limitations}.

A separate benefit of semantic labeling is that it allows the rerendering network to reason about static objects in the scene not captured in the 3D reconstruction, such as lampposts in San Marco Square.
This prevents the network from haphazardly introducing such objects, and instead lets them appear where they are detected in the semantic labeling, which is a significantly simpler task.
In addition, by adding the segmentation labeling to the deep buffer, we allow the appearance encoder to reason about semantic categories like sky or ground when computing the appearance latent vector. 

We compute ``ground truth'' semantic segmentations on the input images $I_i$ using DeepLab~\cite{chen2018deeplab} trained on ADE20K~\cite{zhou2017scene}.
ADE20K contains 150 classes, which we map to a 3-channel color image.
We find that the quality of the semantic labeling is poor on the landmarks themselves, as they contain unique buildings and features, but is reasonable on transient objects.

Using semantic conditioning, the rerendering network takes as input a semantic labeling of the scene.
In order to rerender virtual camera paths, we need to synthesize semantic labelings for each frame in the virtual camera path.
To do so, we train a separate semantic labeling network that takes as input the deep buffer $B_i$, instead of the output image $I_i$, and estimates a ``plausible'' semantic labeling $\hat{S}_i$ for that viewpoint given the rendered deep buffer $B_i$. 
For simplicity, we train a network with the same architecture as the rendering network (minus the injected appearance vector) on samples $(B_i,S_i)$ from the aligned dataset, and we modify the semantic labelings of the ground truth images $S_i$ and mask out the loss on pixels labeled as transient as defined by a curated list of transient object categories in ADE20K.

\section{Evaluation}
\label{sec:evaluation}

\begin{table*}[t]
\small \addtolength{\tabcolsep}{-3.5pt}
\centering
\begin{tabular}{@{\extracolsep{4pt}}lccccccccccccccc}
 & & &\multicolumn{3}{c}{\textbf{I2I}} & \multicolumn{3}{c}{\textbf{+Sem}} & \multicolumn{3}{c}{\textbf{+Sem+BaseApp}} & \multicolumn{3}{c}{\textbf{+Sem+StagedApp}} \\  \cline{4-6} \cline{7-9} \cline{10-12} \cline{13-15}
\textbf{Dataset} & \textbf{\#Images} & \textbf{\#Points} & VGG    & $L_1$  & PSNR  & VGG    & $L_1$  & PSNR   & VGG & $L_1$ & PSNR & VGG & $L_1$ & PSNR \\ \midrule
Sacre Coeur & 1165 & 33M  & 70.78  & 39.98  & 14.36 & 66.17  & 34.78  & 15.62  & \textbf{60.06} & \textbf{21.58} & \textbf{18.98} & 61.23 & 25.22 & 17.81 \\
Trevi & 3006 & 35M  & 86.52  & 42.95  & 14.14 & 81.82  & 36.46  & 15.57  & 79.10 & 28.12 & 17.37 & \textbf{75.55} & \textbf{25.00} & \textbf{18.19} \\
Pantheon & 4972 & 9M  & 68.28  & 39.77  & 14.50 & 67.47  & 36.27  & 15.13  & 64.06 & 28.85 & 16.76 & \textbf{60.66} & \textbf{23.77} & \textbf{17.95} \\
Dubrovnik & 5891 & 33M  & 78.42  & 40.60  & 14.21 & 78.58  & 39.88  & 14.51  & 76.61 & 34.57 & 15.38 & \textbf{71.65} & \textbf{27.48} & \textbf{17.01} \\
San Marco & 7711 & 7M & 80.18  & 44.04  & 13.97 & 78.36  & 39.34  & 14.58  & 70.35 & 26.24 & 17.87 & \textbf{68.96} & \textbf{23.11} & \textbf{18.32} \\ \bottomrule 
\end{tabular}
\caption{Dataset statistics (number of registered images and size of reconstructed point cloud) and average error on the validation set using VGG/perceptual loss (lower is better), $L_1$ loss (lower is better), and PSNR (higher is better), for four methods: an image-to-image baseline (I2I), with semantic conditioning (+Sem), with semantic conditioning and a baseline appearance modeling based on~\cite{zhu2017toward} (+Sem+BaseApp), and with semantic conditioning and staged appearance training (+Sem+StagedApp).\vspace{5pt}}
\label{tab:quantitative_results}
\end{table*}

Here we provide an extensive evaluation of our system.
Please also refer to the supplementary video to best appreciate the quality of the results, available in the project website: {\small\url{https://bit.ly/2UzYlWj}}.

\paragraph{Implementation details}
Our rerendering network is a symmetric encoder-decoder with skip connections, where the
generator is adopted from~\cite{karras2017progressive} without using progressive growing.
We use a multiscale-patchGAN discriminator~\cite{wang2017high} with 3 scales and employ a LSGAN~\cite{mao2017least} loss.
As a reconstruction loss, we use the perceptual loss~\cite{johnson2016perceptual} evaluated at $conv_{i,2}$ for $i \in [1, 5]$ of VGG~\cite{simonyan2014very}.
The appearance encoder architecture is adopted from~\cite{lee2018diverse}, and we use a latent appearance vector $\za \in \mathbb{R}^8$.
We train on 8 GPUs for $\sim 40$ epochs using $256\times256$ crops of input images, but we show compelling results on up to $600 \times 900$ at test time.
The generator runtime for the staged training network is 330 ms for a 512x512 frame on a TitanV without fp16 optimizations.
Architecture and training details can be found in the supplementary material.

\renewcommand{\arraystretch}{0.25}

\begin{figure*}
    \captionsetup[subfigure]{aboveskip=1pt,belowskip=0pt}
    \centering
    \setlength\tabcolsep{1pt}.
        
    \begin{tabular}{ll}
    \rotatebox{90}{\enspace \small  Baseline} & \includegraphics[width=0.97\linewidth]{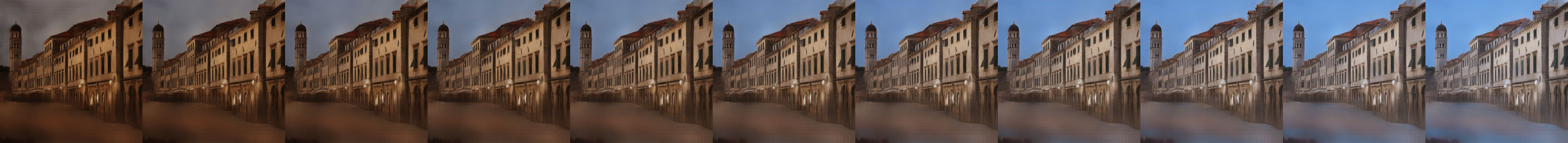} \\ 
    \rotatebox{90}{\enspace \enspace \small  Staged} & \includegraphics[width=0.97\linewidth]{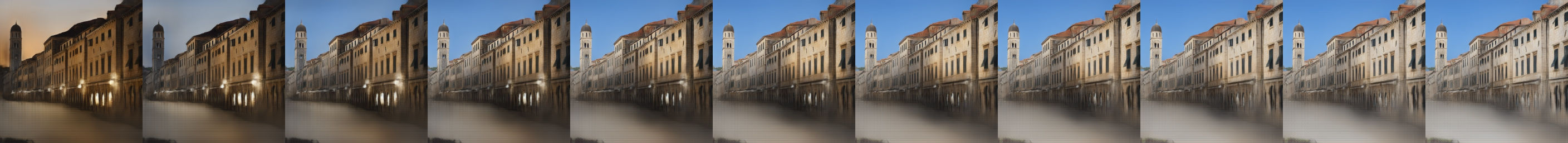} \\
    \rotatebox{90}{\enspace \small Baseline} & \includegraphics[width=0.97\linewidth]{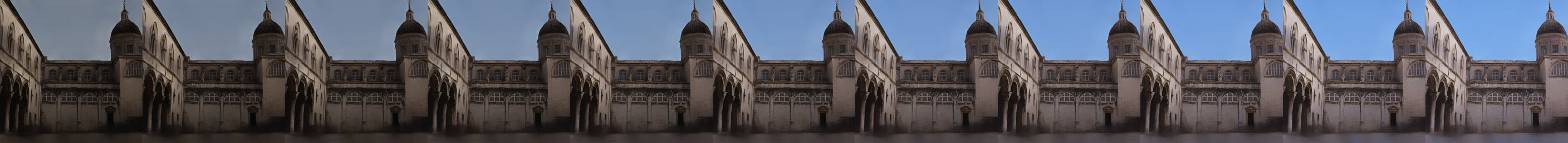} \\ 
    \rotatebox{90}{\enspace \enspace \small  Staged} & \includegraphics[width=0.97\linewidth]{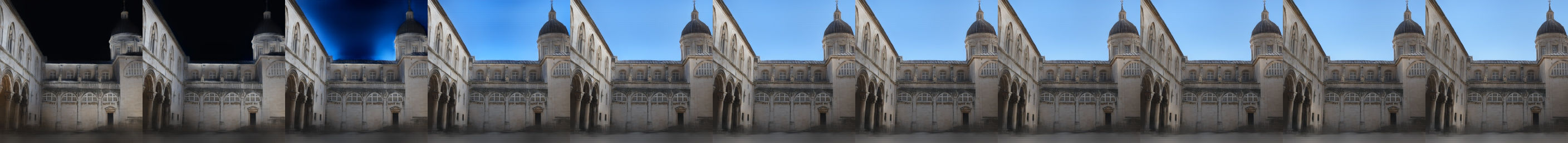} \\
    \end{tabular}
    \caption{Examples of appearance interpolation for a fixed viewpoint. 
    The left- and rightmost appearances are captured from real images, and the intermediate frames are generated by linearly interpolating the appearances in the latent space. 
    Notice how the baseline method is unable to capture complex scenes, like the sunset and night scene, and its interpolations are rather linear, as can be appreciated in the street lamps (top). 
    The staged training method performs better, but generates twilight artifacts in the sky when interpolating between day and night appearances (bottom).}
    \label{fig:appearance_interpolation}
\end{figure*}

\paragraph{Datasets}
We evaluate our method on five datasets reconstructed with COLMAP~\cite{schonberger2016colmap} from public images, summarized in Table~\ref{tab:quantitative_results}. A separate model is trained for each dataset.  
We create aligned datasets by rendering the reconstructed point clouds with a minimum dimension of 600 pixels, and throw away sparse renderings ($>$85$\%$ empty pixels), and small images ($<$450 pixels across).
We randomly select a validation set of 100 images per dataset.

\paragraph{Ablative study}
We perform an ablative study of our system and compare the proposed methods in Figure~\ref{fig:ablative_study}.
The results of the image-to-image translation baseline method contain additional blurry artifacts near the ground because it hallucinates the locations of pedestrians. 
Using semantic conditioning, the results improve slightly in those regions. 
Finally, encoding the appearance of the input photo allows the network to match the appearance.
The staged training recovers a closer appearance in San Marco and Pantheon datasets (two bottom rows). However, in Sacre Coeur (top row), the smallest dataset, the baseline appearance model is able to better capture the general appearance of the image, although the staged training model reproduces the directionality of the lighting with more fidelity.

\paragraph{Reconstruction metrics}
We report image reconstruction errors in the validation set using several metrics: perceptual loss~\cite{johnson2016perceptual}, $L_1$ loss, and PSNR.
We use the ground truth semantic mask from the source image, and we extract the appearance latent vector using the appearance encoder.
Staged training of the appearance fares better than the baseline for all but the smallest dataset (Sacre Coeur), where the staged training overfits to the training data and is unable to generalize. The baseline method assumes a prior distribution of the latent space and is less prone to overfitting at the cost of poorer modeling of appearance.

\begin{figure*}
    \captionsetup[subfigure]{aboveskip=1pt,belowskip=0pt}
    \centering
    \adjincludegraphics[height=2.03cm,trim={0 0 0 0},clip]{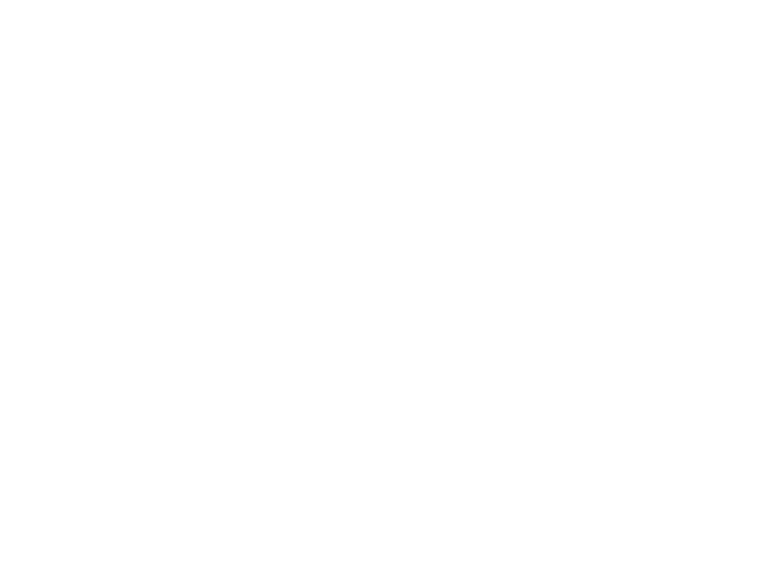}\enspace%
    \adjincludegraphics[height=2.03cm,trim={0 0 0 0},clip]{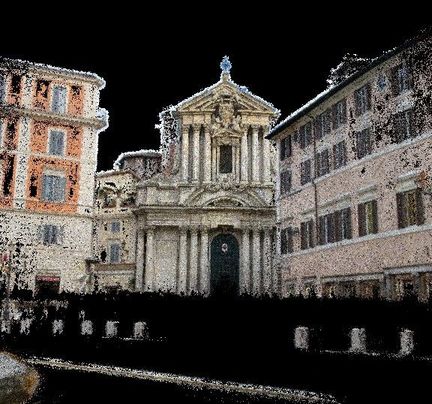}\,%
    \adjincludegraphics[height=2.03cm,trim={0 0 0 0},clip]{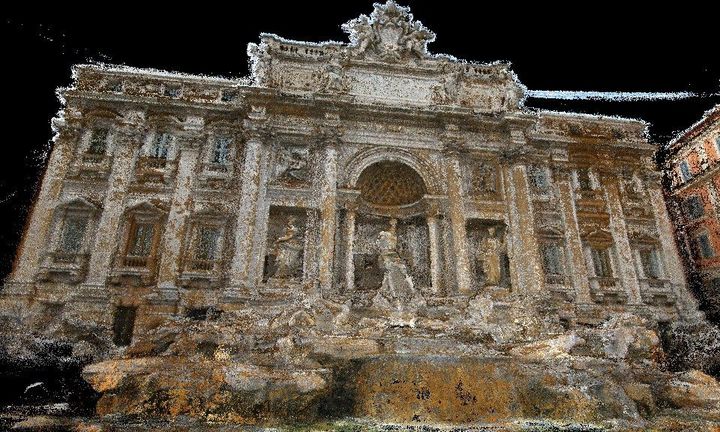}\,%
    \adjincludegraphics[height=2.03cm,trim={0 0 0 0},clip]{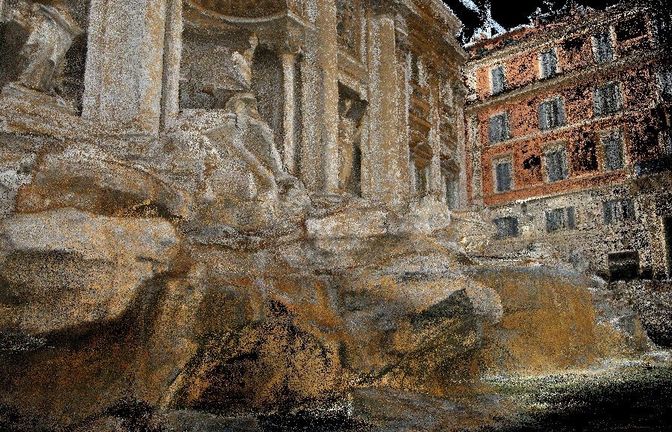}\,%
    \adjincludegraphics[height=2.03cm,trim={0 0 0 0},clip]{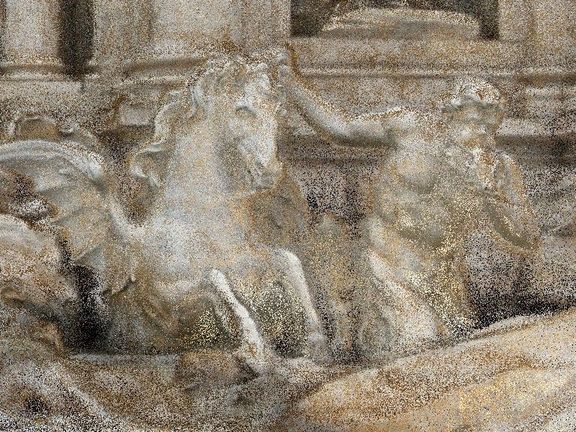}\,%
    \adjincludegraphics[height=2.03cm,trim={0 0 0 0},clip]{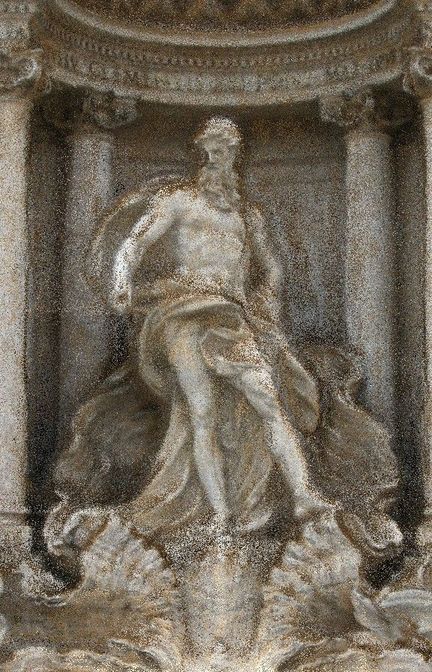}\,%
    \adjincludegraphics[height=2.03cm,trim={0 0 0 0},clip]{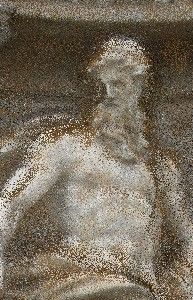}\\[0.04in]
    \adjincludegraphics[height=2.03cm,trim={0 0 0 0},clip]{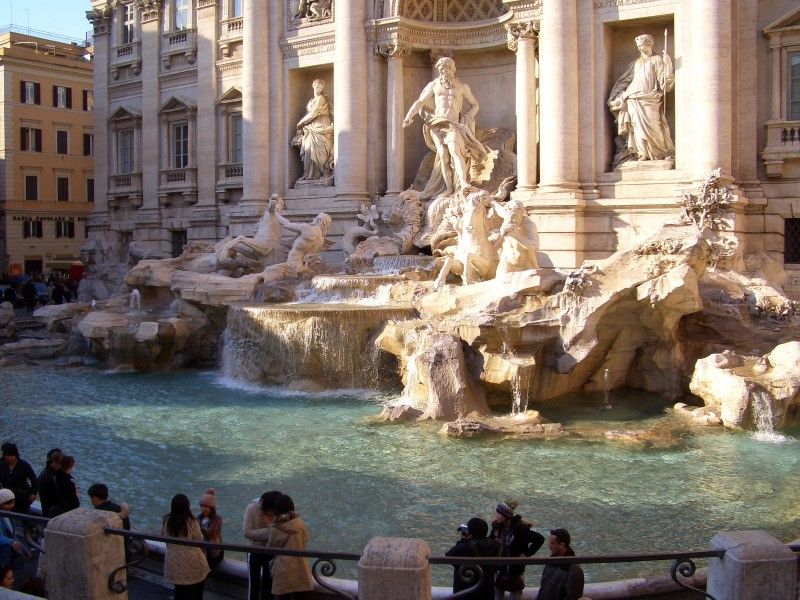}\enspace%
    \adjincludegraphics[height=2.03cm,trim={0 0 {.75\width} 0},clip]{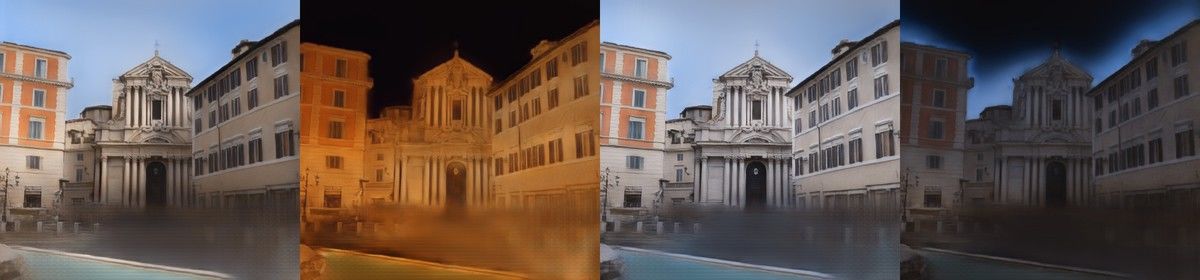}\,%
    \adjincludegraphics[height=2.03cm,trim={0 0 {.75\width} 0},clip]{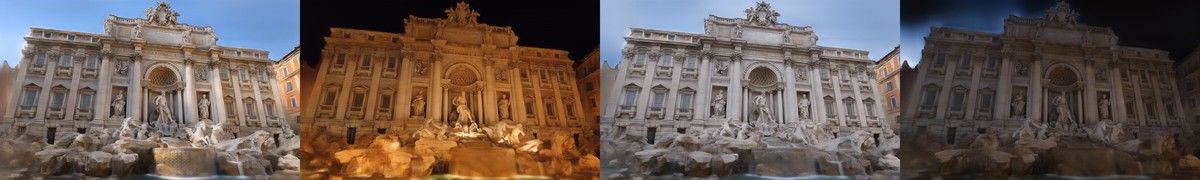}\,%
    \adjincludegraphics[height=2.03cm,trim={0 0 {.75\width} 0},clip]{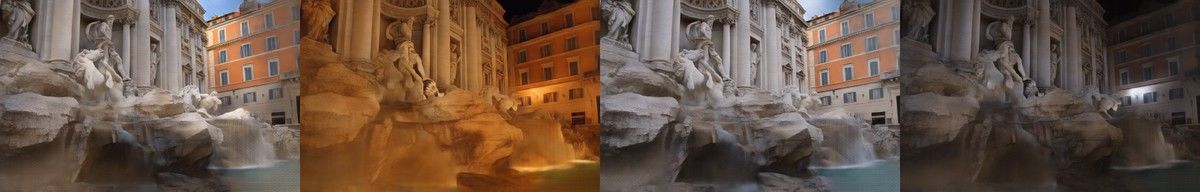}\,%
    \adjincludegraphics[height=2.03cm,trim={0 0 {.75\width} 0},clip]{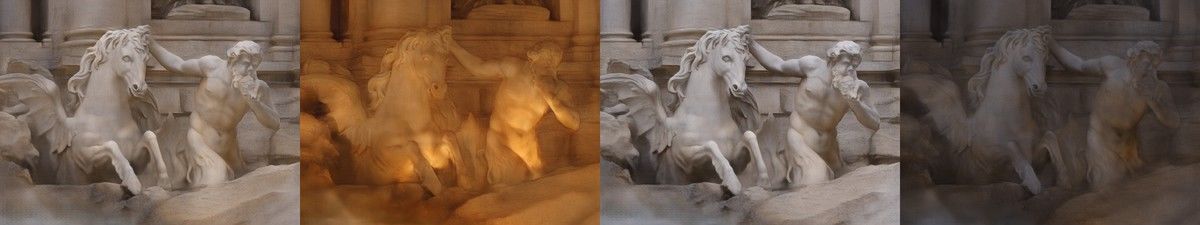}\,%
    \adjincludegraphics[height=2.03cm,trim={0 0 {.75\width} 0},clip]{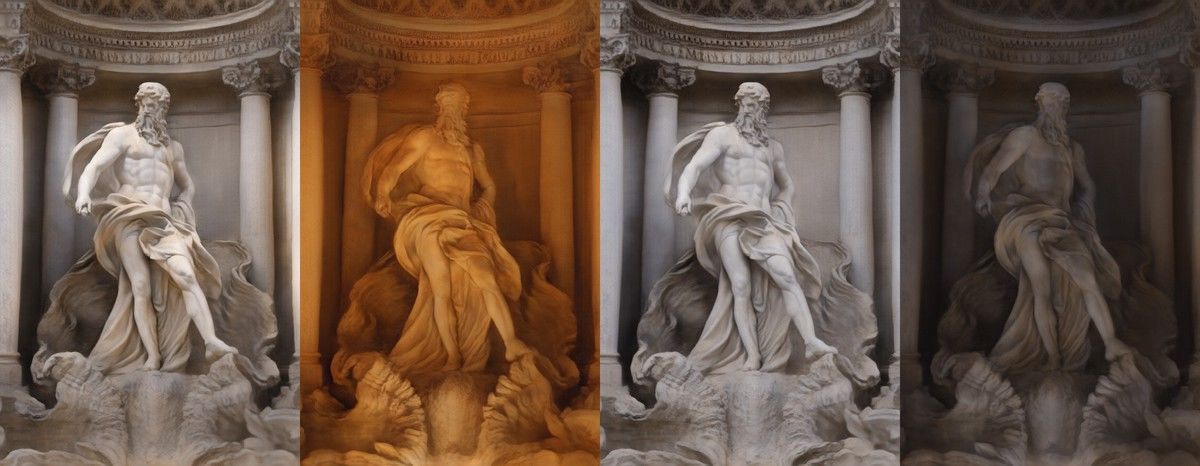}\,%
    \adjincludegraphics[height=2.03cm,trim={0 0 0 0},clip]{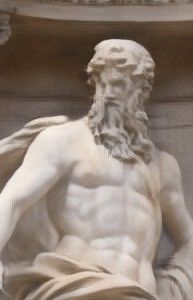}\\
    \adjincludegraphics[height=2.03cm,trim={0 0 0 0},clip]{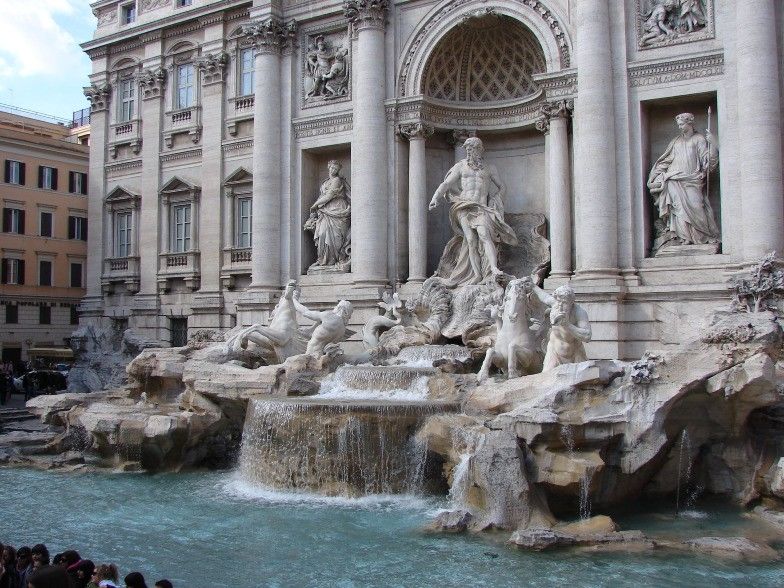}\enspace%
    \adjincludegraphics[height=2.03cm,trim={{.5\width} 0 {.25\width} 0},clip]{figures_opt/appearance_variation/0171_montage_cropped.jpg}\,%
    \adjincludegraphics[height=2.03cm,trim={{.5\width} 0 {.25\width} 0},clip]{figures_opt/appearance_variation/0500_montage.jpg}\,%
    \adjincludegraphics[height=2.03cm,trim={{.5\width} 0 {.25\width} 0},clip]{figures_opt/appearance_variation/0584_montage.jpg}\,%
    \adjincludegraphics[height=2.03cm,trim={{.5\width} 0 {.25\width} 0},clip]{figures_opt/appearance_variation/0796_montage.jpg}\,%
    \adjincludegraphics[height=2.03cm,trim={{.5\width} 0 {.25\width} 0},clip]{figures_opt/appearance_variation/0591_montage.jpg}\,%
    \adjincludegraphics[height=2.03cm,trim={0 0 0 0},clip]{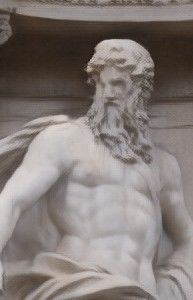}\\
    \adjincludegraphics[height=2.03cm,trim={0 0 0 0},clip]{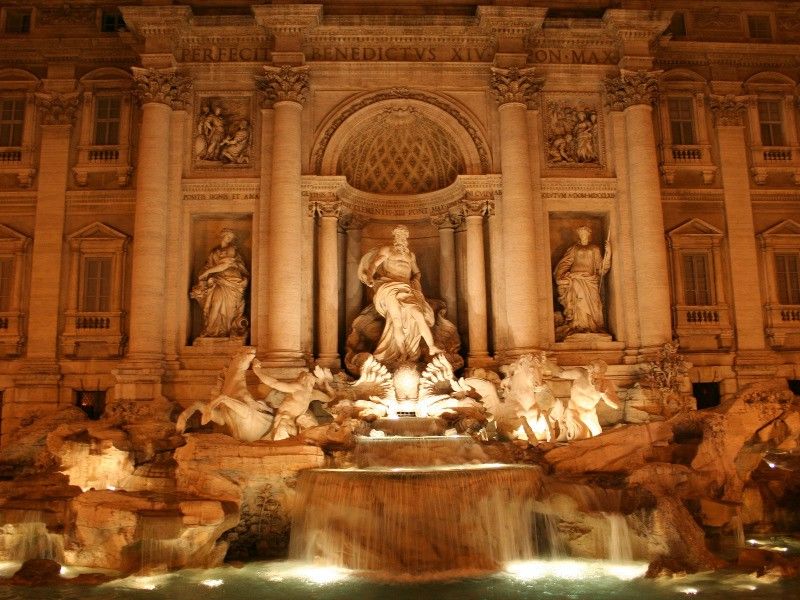}\enspace%
    \adjincludegraphics[height=2.03cm,trim={{.25\width} 0 {.5\width} 0},clip]{figures_opt/appearance_variation/0171_montage_cropped.jpg}\,%
    \adjincludegraphics[height=2.03cm,trim={{.25\width} 0 {.5\width} 0},clip]{figures_opt/appearance_variation/0500_montage.jpg}\,%
    \adjincludegraphics[height=2.03cm,trim={{.25\width} 0 {.5\width} 0},clip]{figures_opt/appearance_variation/0584_montage.jpg}\,%
    \adjincludegraphics[height=2.03cm,trim={{.25\width} 0 {.5\width} 0},clip]{figures_opt/appearance_variation/0796_montage.jpg}\,%
    \adjincludegraphics[height=2.03cm,trim={{.25\width} 0 {.5\width} 0},clip]{figures_opt/appearance_variation/0591_montage.jpg}\,%
    \adjincludegraphics[height=2.03cm,trim={0 0 0 0},clip]{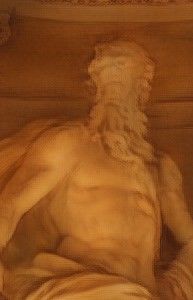}\\
    \adjincludegraphics[height=2.03cm,trim={0 0 0 0},clip]{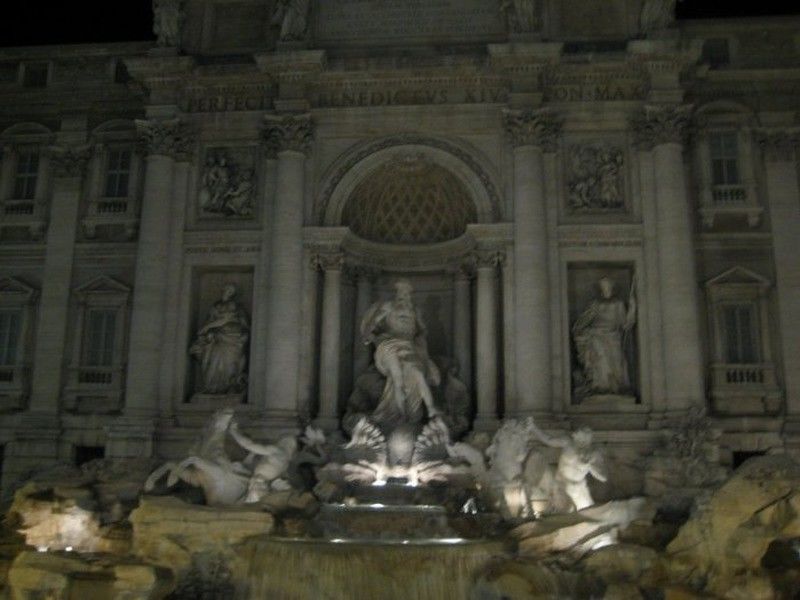}\enspace%
    \adjincludegraphics[height=2.03cm,trim={{.75\width} 0 0 0},clip]{figures_opt/appearance_variation/0171_montage_cropped.jpg}\,%
    \adjincludegraphics[height=2.03cm,trim={{.75\width} 0 0 0},clip]{figures_opt/appearance_variation/0500_montage.jpg}\,%
    \adjincludegraphics[height=2.03cm,trim={{.75\width} 0 0 0},clip]{figures_opt/appearance_variation/0584_montage.jpg}\,%
    \adjincludegraphics[height=2.03cm,trim={{.75\width} 0 0 0},clip]{figures_opt/appearance_variation/0796_montage.jpg}\,%
    \adjincludegraphics[height=2.03cm,trim={{.75\width} 0 0 0},clip]{figures_opt/appearance_variation/0591_montage.jpg}\,%
    \adjincludegraphics[height=2.03cm,trim={0 0 0 0},clip]{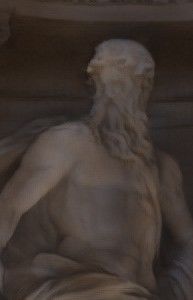}%
    \caption{We capture the appearance of the original images in the left column, and rerender several viewpoints under them.  The last column is a detail of the previous one.
    The top row shows the renderings part of the input to the rerenderer, that exhibit artifacts like incomplete features in the statue, and an inconsistent mix of day and night appearances.
    Note the hallucinated twilight scene in the sky using the last appearance. \footnotesize{Image credits: Flickr users William Warby, Neil Rickards, Rafael Jimenez, acme401 (Creative Commons).} \vspace{-6pt}}
    \label{fig:appearance_variation}
\end{figure*}

\begin{figure*}
    \centering
    \captionsetup[subfigure]{labelformat=empty,aboveskip=1pt,belowskip=0pt}
    \begin{subfigure}[b]{0.120\textwidth}
        \includegraphics[width=\textwidth]{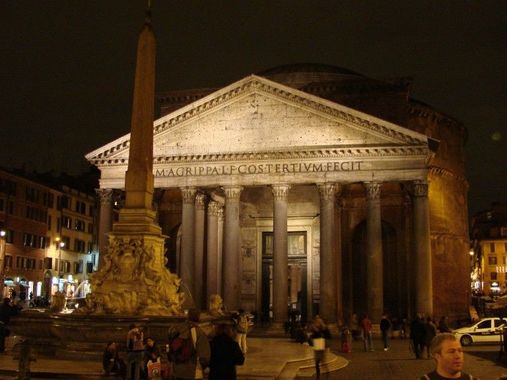}
    \end{subfigure}
    \begin{subfigure}[b]{0.120\textwidth}
        \includegraphics[width=\textwidth]{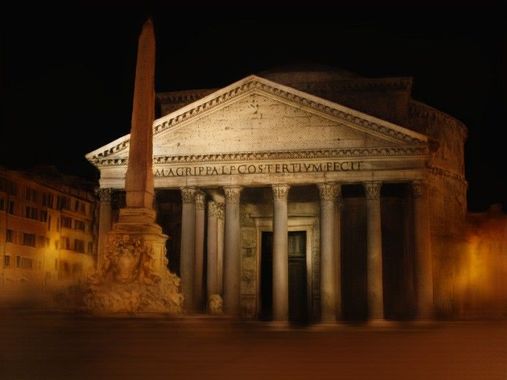}
    \end{subfigure}%
    \begin{subfigure}[b]{0.120\textwidth}
        \caption{}
        \includegraphics[width=\textwidth]{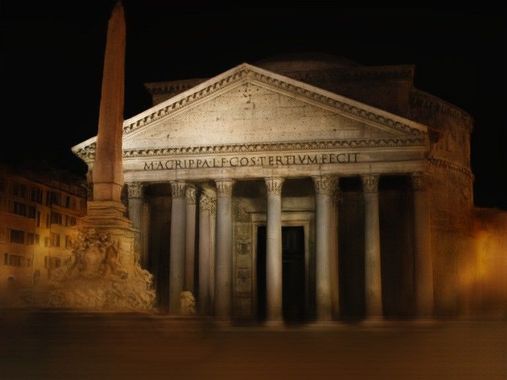}
    \end{subfigure}%
    \begin{subfigure}[b]{0.120\textwidth}
        \caption{}
        \includegraphics[width=\textwidth]{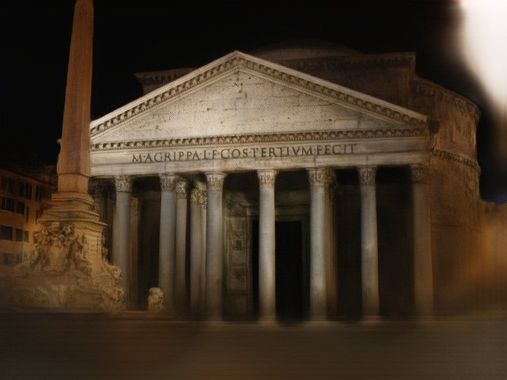}
    \end{subfigure}%
    \begin{subfigure}[b]{0.120\textwidth}
        \caption{}
        \includegraphics[width=\textwidth]{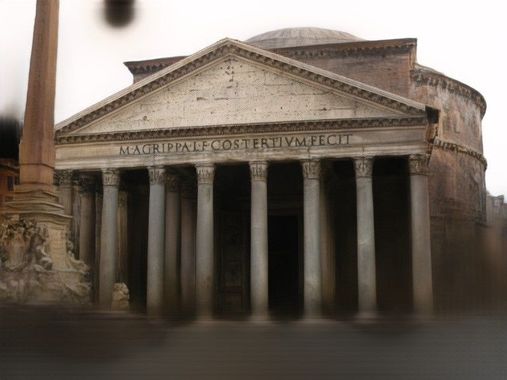}
    \end{subfigure}%
    \begin{subfigure}[b]{0.120\textwidth}
        \caption{}
        \includegraphics[width=\textwidth]{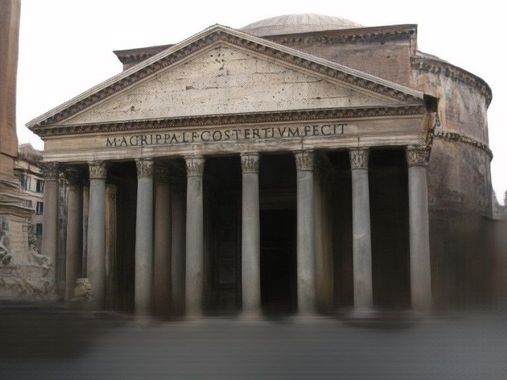}
    \end{subfigure}%
    \begin{subfigure}[b]{0.120\textwidth}
        \includegraphics[width=\textwidth]{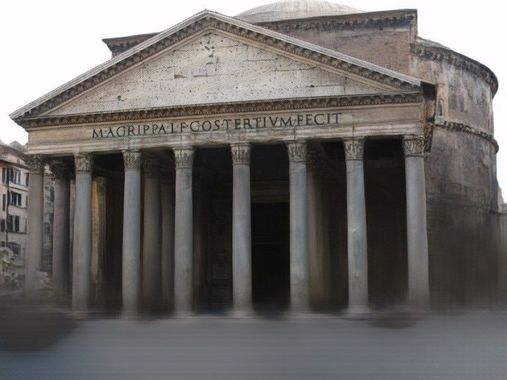}
    \end{subfigure}
    \begin{subfigure}[b]{0.120\textwidth}
        \includegraphics[width=\textwidth]{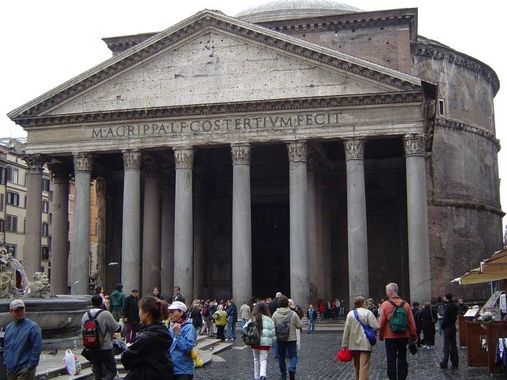}
    \end{subfigure}\\
    \begin{subfigure}[b]{0.120\textwidth}
        \includegraphics[width=\textwidth]{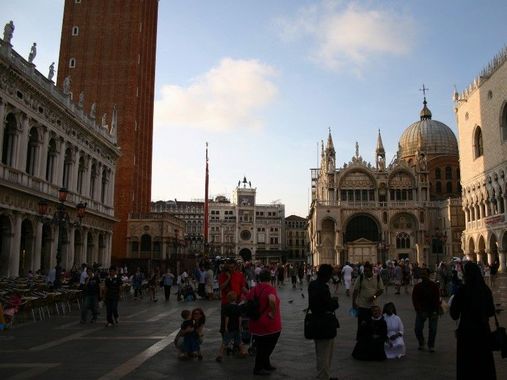}
        \caption{\hspace{-0.5em}Photo}
    \end{subfigure}
    \begin{subfigure}[b]{0.120\textwidth}
        \includegraphics[width=\textwidth]{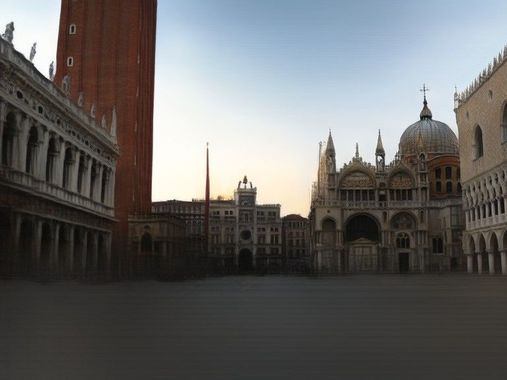}
        \caption{\hspace{-0.5em}Frame 0}
    \end{subfigure}%
    \begin{subfigure}[b]{0.120\textwidth}
        \includegraphics[width=\textwidth]{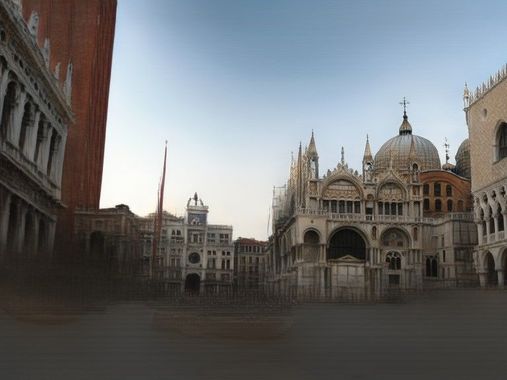}
        \caption{\hspace{-0.5em}Frame 20}
    \end{subfigure}%
    \begin{subfigure}[b]{0.120\textwidth}
        \includegraphics[width=\textwidth]{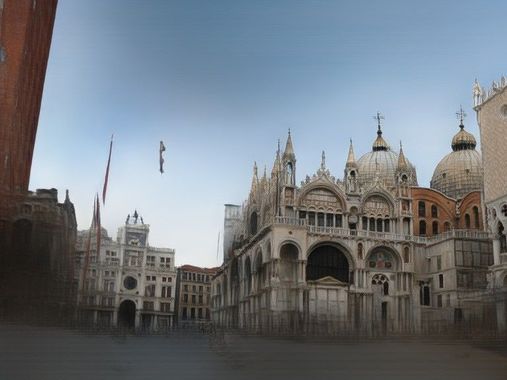}
        \caption{\hspace{-0.5em}Frame 40}
    \end{subfigure}%
    \begin{subfigure}[b]{0.120\textwidth}
        \includegraphics[width=\textwidth]{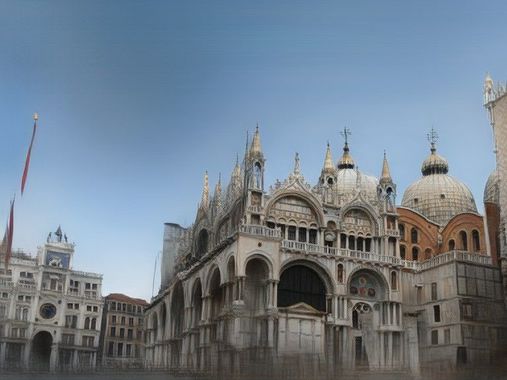}
        \caption{\hspace{-0.5em}Frame 60}
    \end{subfigure}%
    \begin{subfigure}[b]{0.120\textwidth}
        \includegraphics[width=\textwidth]{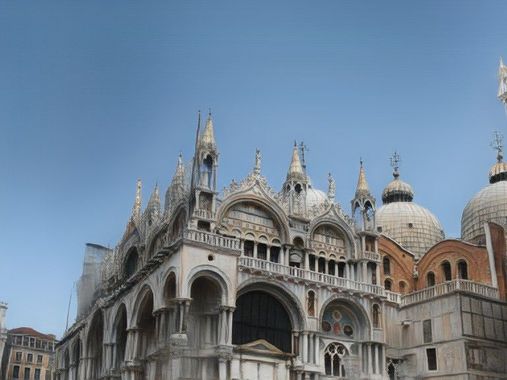}
        \caption{\hspace{-0.5em}Frame 80}
    \end{subfigure}%
    \begin{subfigure}[b]{0.120\textwidth}
        \includegraphics[width=\textwidth]{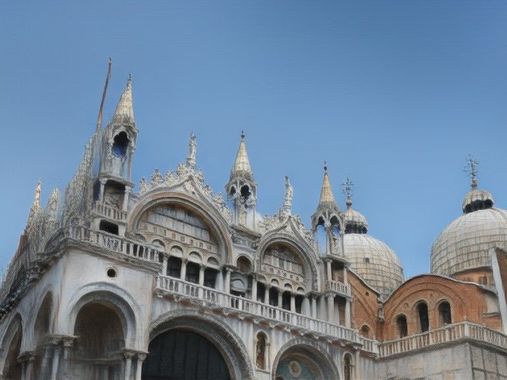}
        \caption{\hspace{-0.5em}Frame 100}
    \end{subfigure}
    \begin{subfigure}[b]{0.120\textwidth}
        \includegraphics[width=\textwidth]{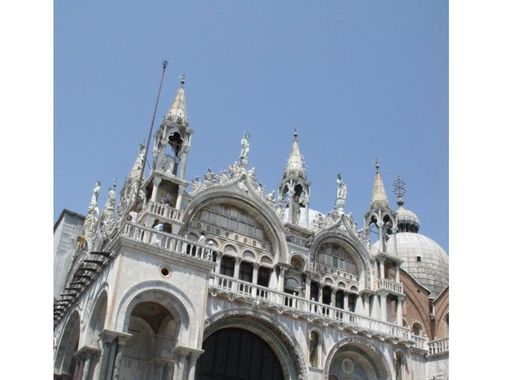}
        \caption{\hspace{-0.5em}Photo}
    \end{subfigure}
    \caption{Frames from a synthesized camera path that smoothly transitions from the photo on the left to the photo on the right by smoothly interpolating both viewpoint and the latent appearance vectors.
    Please see the supplementary video. \footnotesize{Photo Credits: Allie Caulfield, Tahbepet, Till Westermayer, Elliott Brown (Creative Commons).} \vspace{-4pt}}
    \label{fig:image_interpolations}
\end{figure*}

\paragraph{Appearance interpolation}
The rerendering network allows for interpolating the appearance of two images by interpolating their latent appearance vectors. Figure~\ref{fig:appearance_interpolation} depicts two examples, showing that the staged training approach is able to generate more complex appearance changes, although its generated interpolations lack realism when transitioning between day and night. In the following, we only show results for the staged training model. 

\paragraph{Appearance transfer}
Figure~\ref{fig:appearance_variation} demonstrates how our full model can transfer the appearance of a given photo to others.  It shows realistic renderings of the Trevi fountain from five different viewpoints under four different appearances obtained from other photos. 
Note the sunny highlights and the spotlit night illumination appearance of the statues.
However, these details can flicker when synthesizing a smooth camera path or smoothly interpolating the appearance in the latent space, as seen in the supplementary video.

\paragraph{Image interpolation}
Figure~\ref{fig:image_interpolations} shows sets of two images and frames of smooth image interpolations between them, where both viewpoint and appearance transition smoothly between them. Note how the illumination of the scene can transition smoothly from night to day. The quality of the results is best appreciated in the supplementary video.
\begin{figure}
    \vspace*{10pt}
    \centering
    \captionsetup[subfigure]{aboveskip=1pt,belowskip=0pt}
    \begin{subfigure}[b]{0.49\linewidth}
        \includegraphics[width=0.49\linewidth]{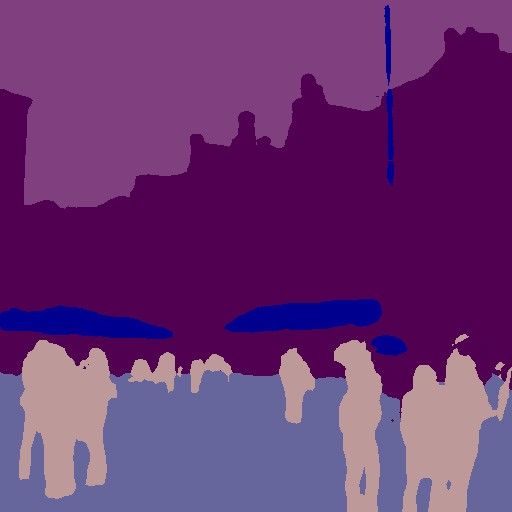}
        \includegraphics[width=0.49\linewidth]{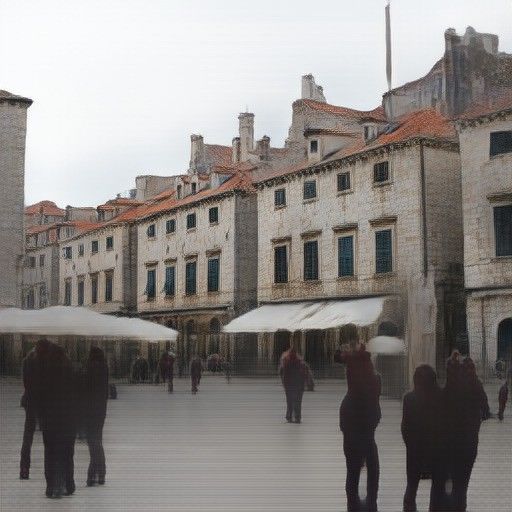}
        \caption{w/ GT segmentation}
    \end{subfigure}
    \begin{subfigure}[b]{0.49\linewidth}
        \includegraphics[width=0.49\linewidth]{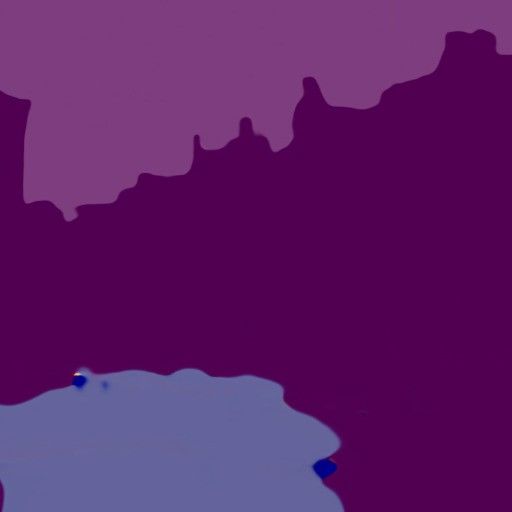}
        \includegraphics[width=0.49\linewidth]{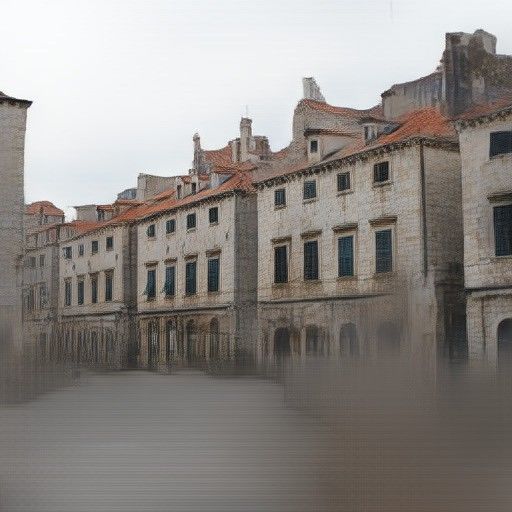}
        \caption{w/ predicted segmentations}
        \label{fig:bad_segmention}
    \end{subfigure}\\
    \caption{Example semantic labelings and output renders when using the ``ground truth'' segmentation mask computed from the corresponding real image (from the validation set) and the predicted one from the associated deep buffer.
    Note the artifacts on the bottom right where the ground is misclassified as building.}
    \label{fig:semantic_consistency}
    \vspace*{10pt}
\end{figure}

\paragraph{Semantic consistency}
Figure~\ref{fig:semantic_consistency} shows the output of the staged training model with ground truth and predicted segmentation masks.
Using the predicted masks, the network produces similar results on the building and renders a scene free of people. 
Note however how the network depicts pedestrians as black, ghostly figures when they appear in the segmentation mask.

\paragraph{Comparison to 3D reconstruction methods}

\begin{figure}[t!]

    \captionsetup[subfigure]{aboveskip=1pt,belowskip=1pt}
    \centering
    \begin{subfigure}[b]{0.325\linewidth}
        \centering
        \includegraphics[width=\linewidth]{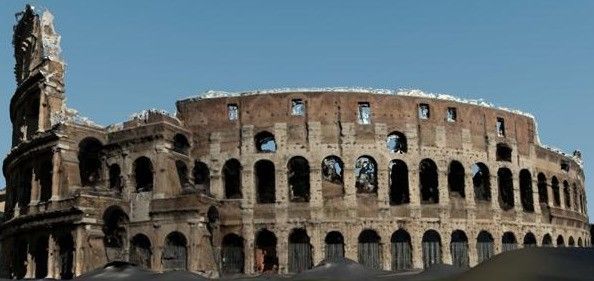}\\
        \includegraphics[width=\linewidth]{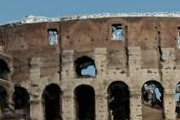}\\
        \includegraphics[width=\linewidth]{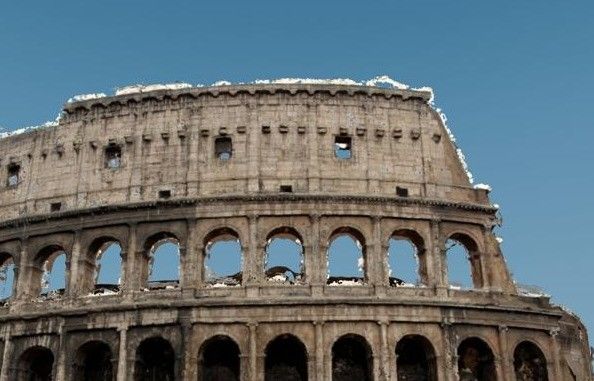}\\
        \includegraphics[width=\linewidth]{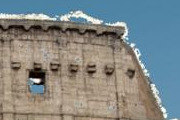}
        \caption{\footnotesize \cite{shan2013visual}}
    \end{subfigure}\,%
    \begin{subfigure}[b]{0.325\linewidth}
        \centering
        \includegraphics[width=\linewidth]{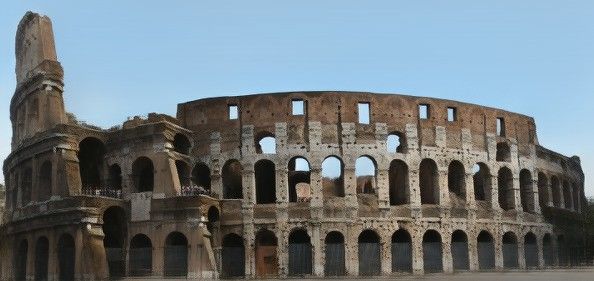}\\
        \includegraphics[width=\linewidth]{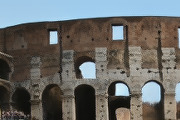}\\
        \includegraphics[width=\linewidth]{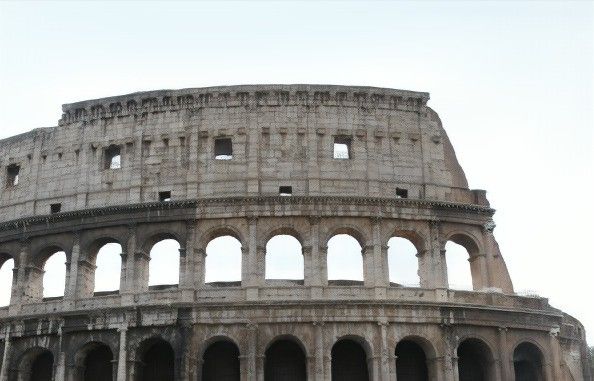}\\
        \includegraphics[width=\linewidth]{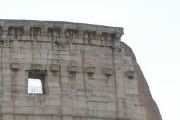}
        \caption{\footnotesize ours}
    \end{subfigure}\,%
    \begin{subfigure}[b]{0.325\linewidth}
        \centering
        \includegraphics[width=\linewidth]{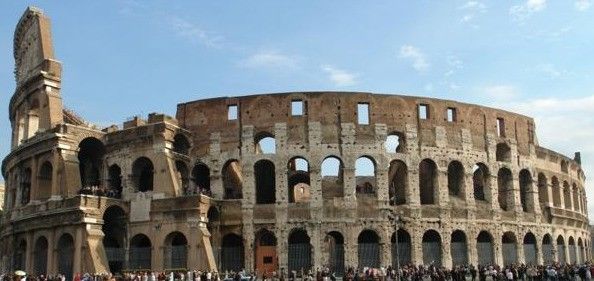}\\
        \includegraphics[width=\linewidth]{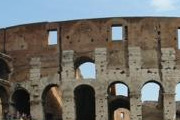}\\
        \includegraphics[width=\linewidth]{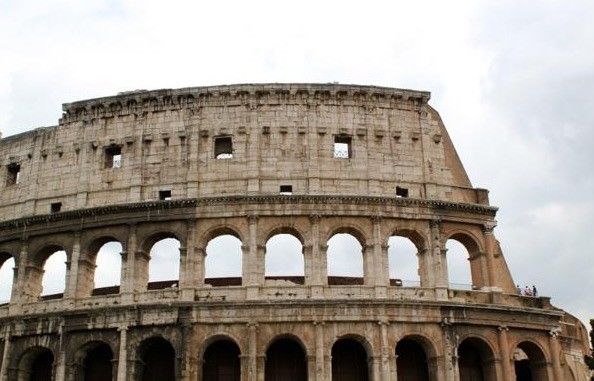}\\
        \includegraphics[width=\linewidth]{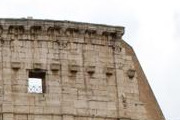}
        \caption{\footnotesize original images}
    \end{subfigure}\\

    \caption{Comparison of~\cite{shan2013visual} and our approach. Rows 1 \& 3: original photos. Rows 2 \& 4: detailed crops.
    \footnotesize{Image credits: Graeme Churchard, Sarah-Rose (Creative Commons).}
    }
    \vspace*{10pt}
    \label{fig:visual_turing_test}
\end{figure}

We evaluated our technique against the one of Shan \etal~\cite{shan2013visual} on the Colosseum, which contains 3K images, 10M color vertices and 48M triangles and was generated from Flickr, Google Street View, and aerial images.
Their 3D representation is a dense vertex-colored mesh, where the albedo and vertex normals are jointly recovered together with a simple 8-dimensional lighting model (diffuse, plus directional lighting) for each image in the photo collection.

Figure~\ref{fig:visual_turing_test} compares both methods and the original ground truth image.
Their method suffers from floating white geometry on the top edge of the Colosseum, and has less detail, although it recovers the lighting better than our method, thanks to its explicit lighting reasoning.
Note that both models are accessing the test image to compute lighting coefficients and appearance latent vectors, with dimension 8 in both cases, and that we use the predicted segmentation labelings from $B_i$.

We ran a randomized user study on 20 random sets of output images that do not contain close-ups of people or cars, and were not in our training set. 
For each viewpoint, 200 participants chose ``which image looks most real?'' between an output of their system and ours (without seeing the original).
Respondents preferred images generated by our system a $69.9\%$ of the time, with our technique being preferred on all but one of the images.
We show the 20 random sets of the user study in the supplementary material.

\section{Discussion}
\label{sec:discussion}

\begin{figure}
    \centering
    \captionsetup[subfigure]{aboveskip=0pt,belowskip=0pt}
    \begin{subfigure}[b]{\linewidth}
        \includegraphics[width=0.49\linewidth]{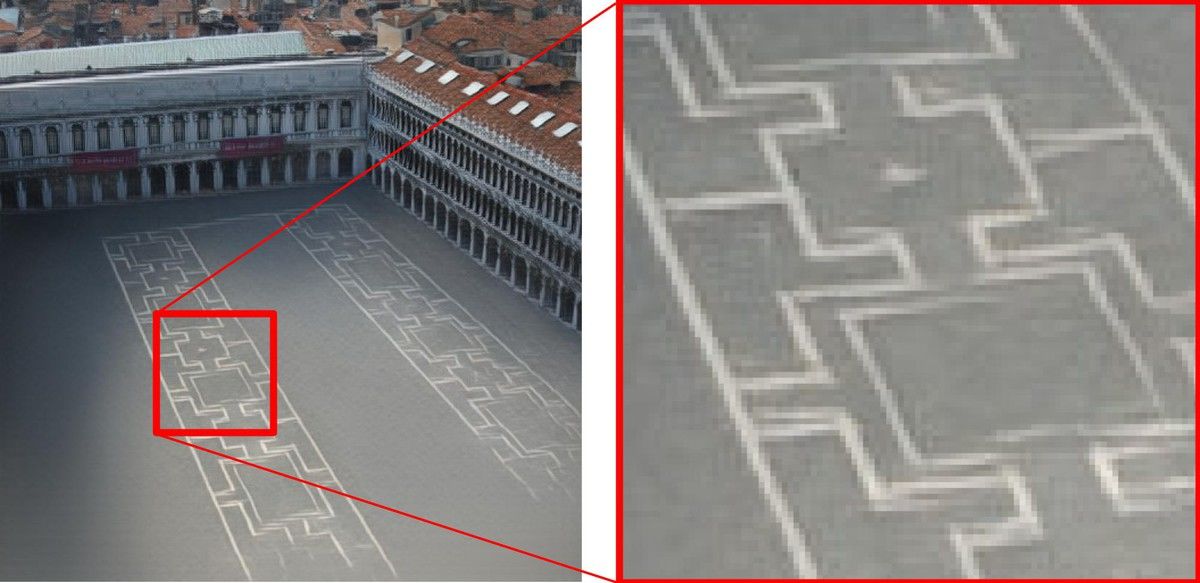}
        \includegraphics[width=0.49\linewidth]{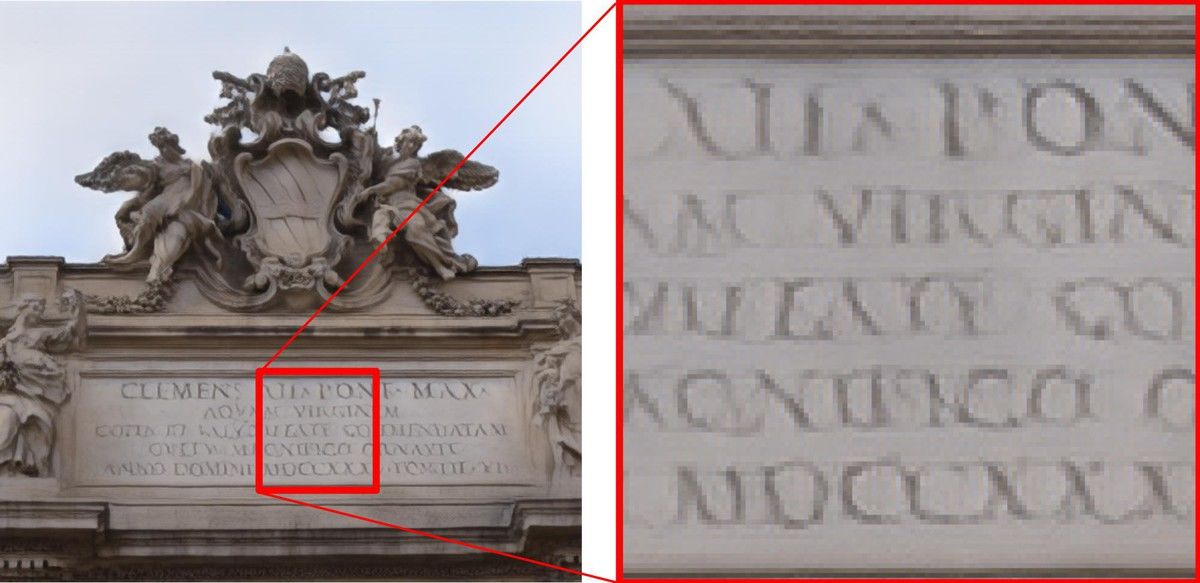}
        \caption{\footnotesize Neural artifacts}
        \label{fig:painterly_artifacts}
    \end{subfigure}\\
    \begin{subfigure}[b]{0.49\linewidth}
        \includegraphics[width=0.49\linewidth]{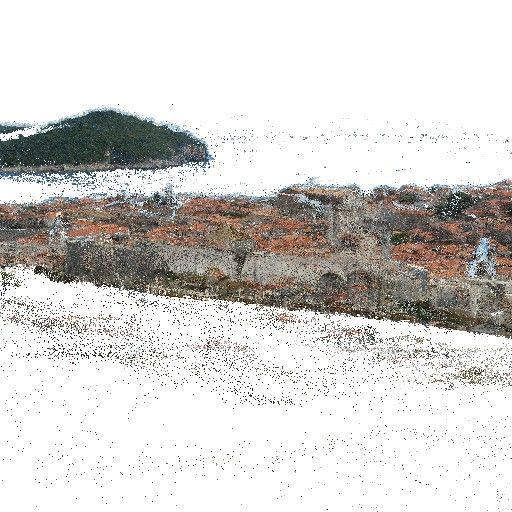}\hspace{2pt}%
        \includegraphics[width=0.49\linewidth]{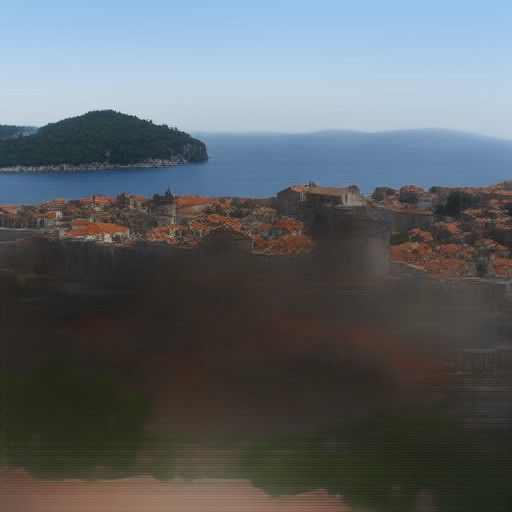}
        \caption{Sparse reconstructions}
        \label{fig:sparse_reconstructions}
    \end{subfigure}\hspace{3pt}%
    \begin{subfigure}[b]{0.49\linewidth}
        \includegraphics[width=0.49\linewidth]{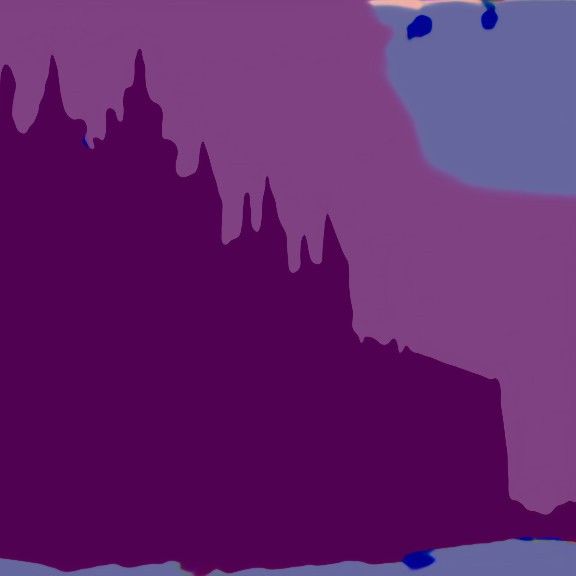}\hspace{2pt}%
        \includegraphics[width=0.49\linewidth]{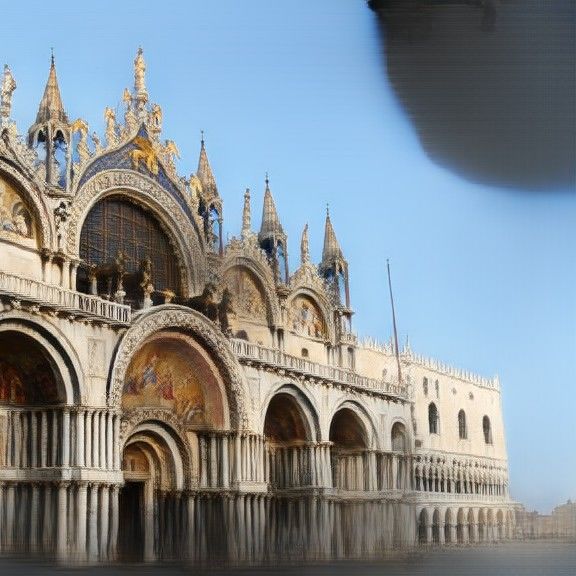}
        \caption{\footnotesize Segmentations artifacts}
        \label{fig:segmentation_artifacts}
    \end{subfigure}%
    \caption{Limitations of the current system.}
    \vspace*{0pt}
    \label{fig:failure_cases}
\end{figure}

Our system's limitations are significantly different from those of traditional 3D reconstruction pipelines:

\paragraph{Segmentation}
Our model relies heavily on the segmentation mask to synthesize parts of the image not modeled in the proxy geometry, like the ground or sky regions.
Thus our results are very sensitive to errors in the segmentation network, like in the sky region in Figure~\ref{fig:segmentation_artifacts} or an appearing ``ghost pole'' artifact in San Marco (frame 40 of bottom row in Figure~\ref{fig:image_interpolations}, best seen in video).
Jointly training the neural rerenderer together with the segmentation network could reduce such artifacts.\vspace{-1.5pt}

\paragraph{Neural artifacts} 
Neural networks are known to produce screendoor patterns~\cite{odena2016deconvolution} and other intriguing artifacts~\cite{mordvintsev2015inceptionism}.
We observe such artifacts in repeated structures, like the patterns on the floor of San Marco, which in our renderings are misaligned as if hand-painted.
Similarly, the inscription above the Trevi fountain is reproduced with a distorted font (see Figure~\ref{fig:painterly_artifacts}).\vspace{-1.5pt}

\paragraph{Incomplete reconstructions}
Sometimes an image contains partially reconstructed parts of the 3D model, creating large holes in the rendered $B_i$.
This forces the network to hallucinate the incomplete regions, generally leading to blurry outputs (see Figure~\ref{fig:sparse_reconstructions}).\vspace{-1.5pt}

\paragraph{Temporal artifacts} 
When smoothly varying the viewpoint, sometimes the appearance of the scene can flicker considerably, especially under complex appearance, such as when the sun hits the Trevi Fountain, creating complex highlights and cast shadows.
Please see the supplementary video for an example.\vspace{-1.5pt}

\medskip
In summary, we present a first attempt at solving the total scene capture problem.
Using unstructured internet photos, we can train a neural rerendering network that is able to produce highly realistic scenes under different illumination conditions.
We propose a novel staged training approach that better captures the appearance of the scene as seen in internet photos.
Finally, we evaluate our system on five challenging datasets and against state-of-the-art 3D reconstruction methods.

\medskip
\textbf{Acknowledgements:}
We thank Gregory Blascovich for his help in conducting the user study, and Johannes Sch\"{o}nberger and True Price for their help generating datasets.

\clearpage \appendix \section{Supplementary Results}
\begin{figure*}
    \centering
    \includegraphics[width=0.98\linewidth]{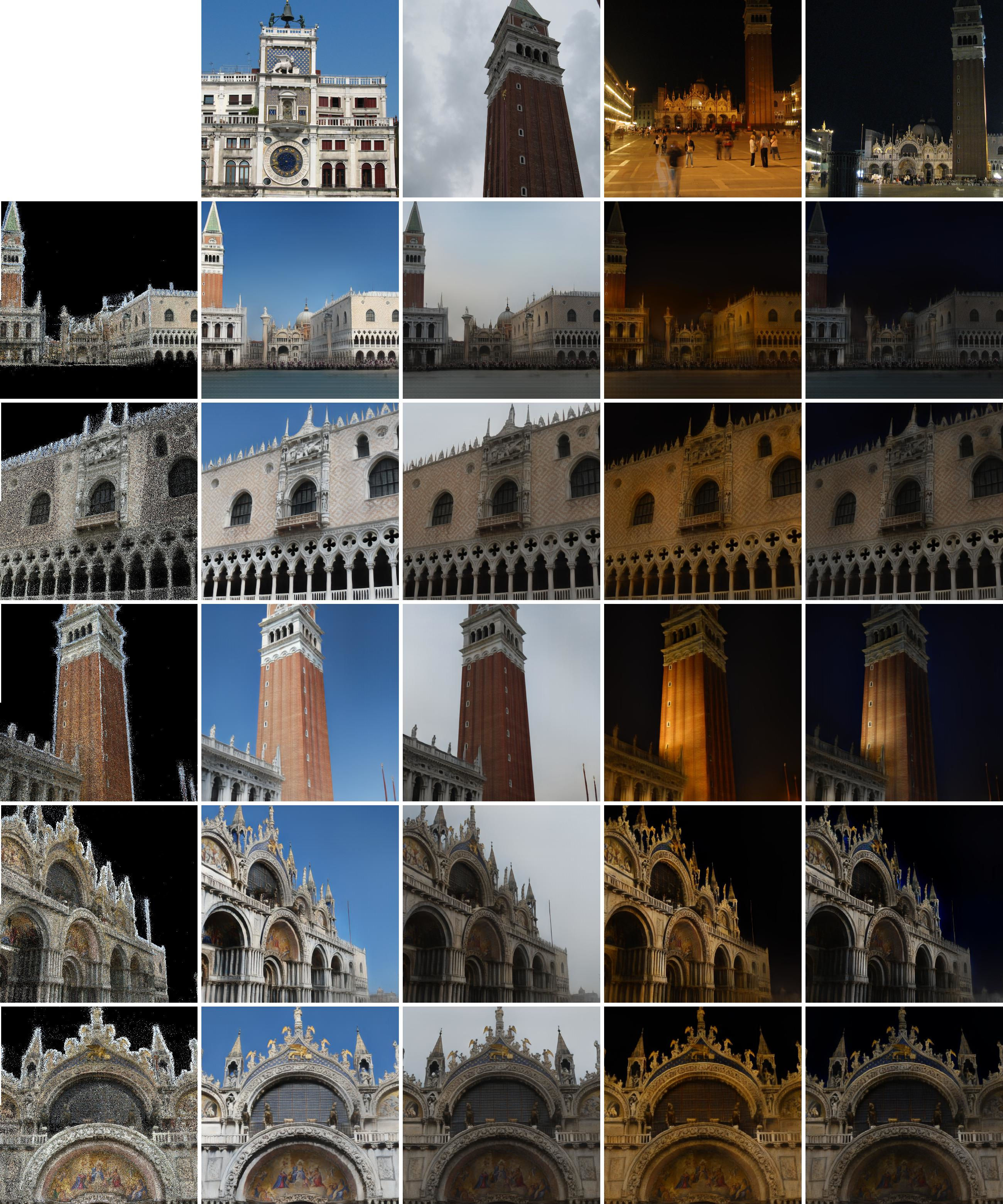}
    \caption{We capture the appearance of the original images in the first row, and rerender several viewpoints under them. The first column shows the rendered point cloud images used as input to the rerenderer. \footnotesize{Image credits: Michael Pate, Jeremy Thompson, Patrick Denker, Rob Young (Creative Commons).}}
    \label{fig:sm-app}
\end{figure*}

\paragraph{Appearance variation.}
Figure~\ref{fig:sm-app} shows additional results of diverse appearances modeled by our proposed staged training method on the San Marco dataset.  As in Figure~\ref{fig:appearance_variation}, it shows realistic renderings of five different scenes/viewpoints under four different appearances obtained from other photos.

\begin{figure*}[p!]
    \centering
    \includegraphics[width=0.98\linewidth]{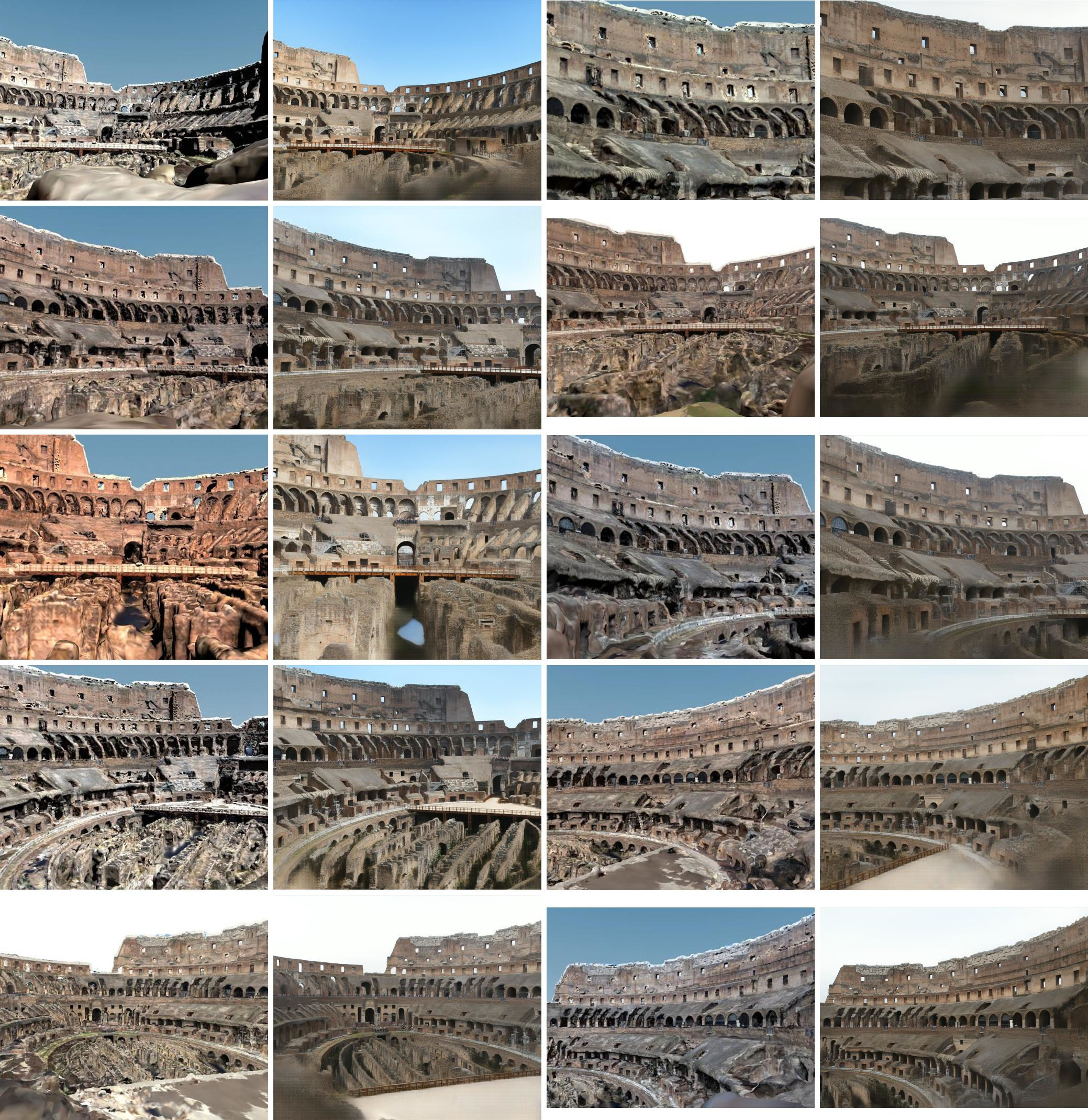}
    \caption{Comparison with Shan \etal~\cite{shan2013visual} -- set 1 of 2. First and third columns show the result of Shan \etal~\cite{shan2013visual}. Second and fourth columns show our result.}
    \label{fig:vtt1}
\end{figure*}

\begin{figure*}[p!]
    \centering
    \includegraphics[width=0.98\linewidth]{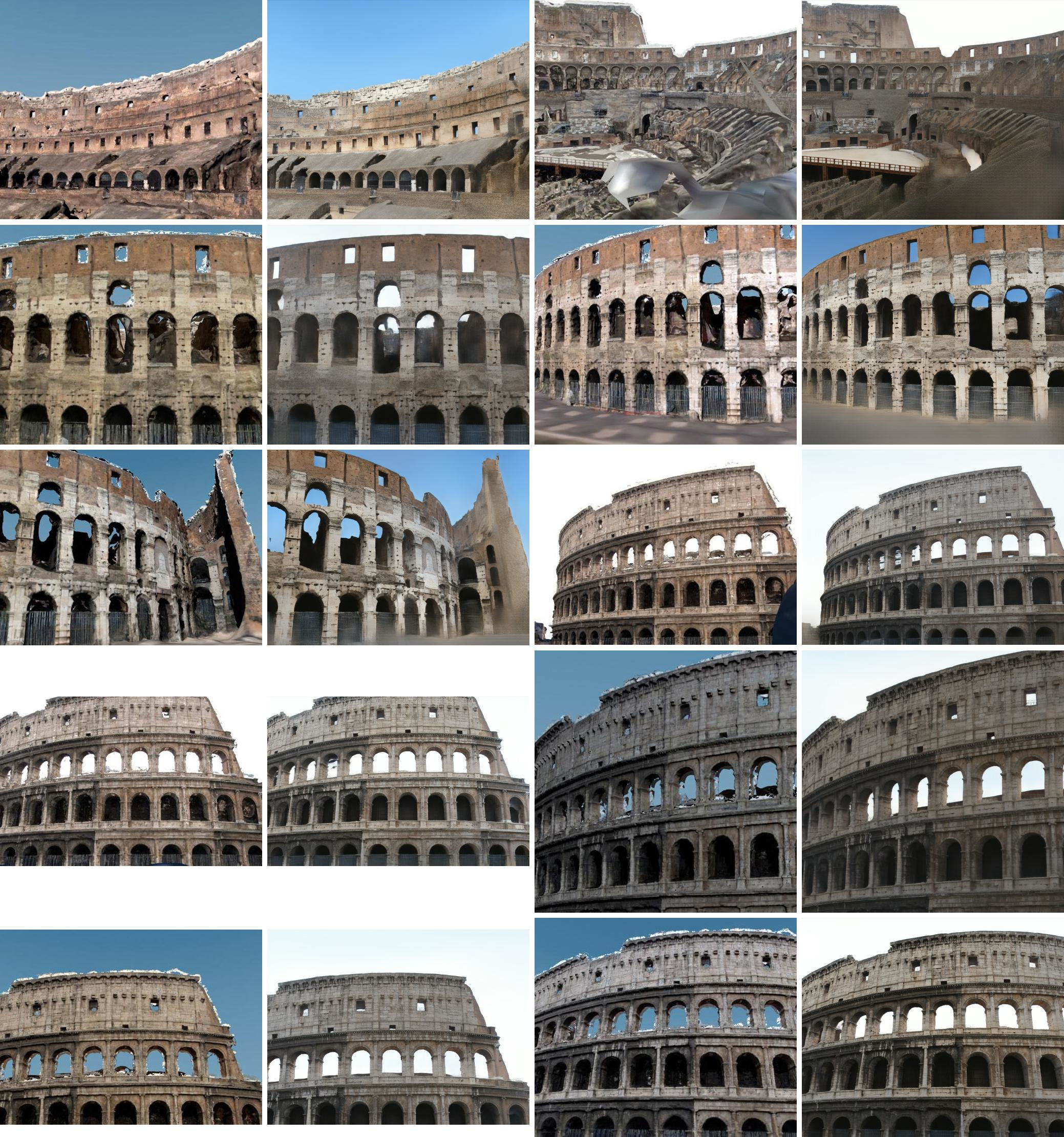}
    \caption{Comparison with Shan \etal~\cite{shan2013visual} -- set 2 of 2. First and third columns show the result of Shan \etal~\cite{shan2013visual}. Second and fourth columns show our result.}
    \label{fig:vtt2}
\end{figure*}

\paragraph{Qualitative comparison.}
We evaluate our technique against Shan \etal~\cite{shan2013visual} on the Colosseum. In Section 4, we report the result of a user study run on 20 randomly selected sets of output images that do not contain close-ups of people or cars, and were not in our training set. 
Figures~\ref{fig:vtt1},~\ref{fig:vtt2} show a side-by-side comparison of all 20 images used in the user study.

\paragraph{Quantitative evaluation with learned segmentations}
To quantitatively evaluate rerendering using estimated segmentation masks, we generate semantic labelings for the validation set, as described in Section~\ref{sec:semantic_conditioning},
and recompute the quantitative metrics, as in Table~\ref{tab:quantitative_results}, for our proposed method. 
Note that estimated semantic maps will not perfectly match those of the ground truth validation images. For example, ground truth semantic maps could contain the segmentation of transient objects, like people or trees. So, it is not fair to compare reconstructions based on estimated segmentation maps to the ground truth validation images.
While results in Table~\ref{tab:quantitative_results_learned_seg} show some performance drop as expected, we still get a reasonable performance compared to that in Table~\ref{tab:quantitative_results}.
In fact, we still perform better than the BicycleGAN baseline on the Trevi, Pantheon and Dubrobnik datasets, even though the BicycleGAN baseline uses ground truth segmentation masks.

\begin{table}[ht!]
\centering
\setlength\extrarowheight{32pt}
\begin{tabular}{lccc}
 & \multicolumn{3}{c}{\textbf{+Sem+StagedApp}} \\ \cline{2-4}
\textbf{Dataset} & VGG & $L_1$ & PSNR \\ \midrule
Sacre Coeur & 67.74 & 28.66 & 16.45  \\
Trevi & 77.35 & 26.03 & 17.90  \\
Pantheon & 62.54 & 25.40 & 17.29 \\
Dubrovnik & 74.44 & 30.39 & 16.18 \\
San Marco & 75.58 & 26.69 & 17.34 \\
\end{tabular}
\caption{We evaluate our staged training approach using estimated segmentation masks, as opposed to Table~\ref{tab:quantitative_results} which uses segmentation masks computed from ground truth validation images.}
\label{tab:quantitative_results_learned_seg}
\end{table}

\section{Implementation Details}
We use different networks for the staged training and the baseline mode. 
We obtain best results for each model with different networks.
Below, we provide an overview of the different architectures used in the staged training and the baseline models.
Code will be available at {\small\url{https://bit.ly/2UzYlWj}}.

\subsection{Neural rerender network architecture}
Our rerendering network is a symmetric encoder-decoder with skip connections.
The generator is adopted from~\cite{karras2017progressive} without using progressive growing.
Specifically, we extend the GAN architecture in~\cite{karras2017progressive} to a conditional GAN setting. 
The encoder/decoder operates at a $256 \times 256$ resolution, with 6 downsampling/upsampling blocks.
Each block has a downsampling/upsampling layer followed by two single-strided $3 \times 3$ \emph{conv} layers with a \emph{leaky ReLu} ( $\alpha=0.2$) and \emph{pixel-norm}~\cite{karras2017progressive} layers.
We add skip connections between the encoder and decoder by concatenating feature maps at the beginning of each decoder block.
We use 64 feature maps at the first encoder and double the size of feature maps after each downsampling layer until it reaches size 512.

\subsection{Appearance encoder architecture}
We implement the appearance encoder architecture used in~\cite{lee2018diverse} except that we add \emph{pixel-norm}~\cite{karras2017progressive} layers after each downsampling block. We observe that adding
a pixel-wise normalization layer stabilizes the training while at the same time avoids mixing information between different pixels as in \emph{instance norm} or \emph{batch norm}. 
We use a latent appearance vector $\za \in \mathbb{R}^8$. The latent vector is injected at the bottleneck between the encoder and decoder in the rendering network. We tile $\za$ to match the dimension of feature maps at the bottleneck and concatenate it to the feature maps channel-wise.

\begin{figure}[t!]
    \centering
    \captionsetup[subfigure]{aboveskip=1pt,belowskip=2pt}
    \begin{subfigure}[b]{0.32\linewidth}
        \centering
        \includegraphics[width=\linewidth]{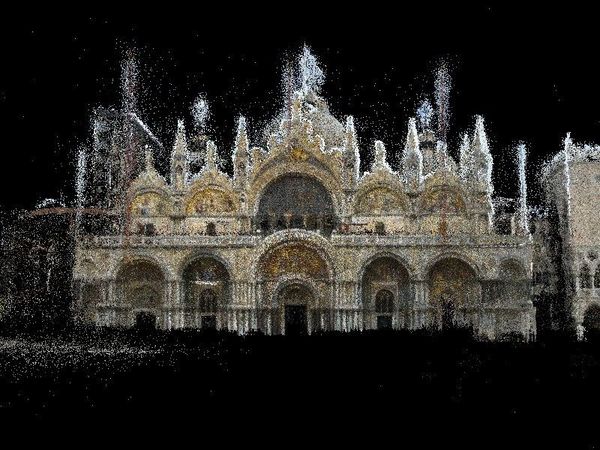}\\
        \includegraphics[width=\linewidth]{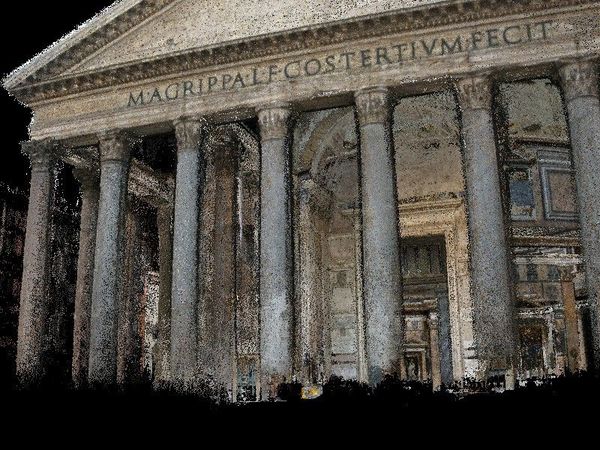}
        \caption{initial rendering}
    \end{subfigure}\,%
    \begin{subfigure}[b]{0.32\linewidth}
        \centering
        \includegraphics[width=\linewidth]{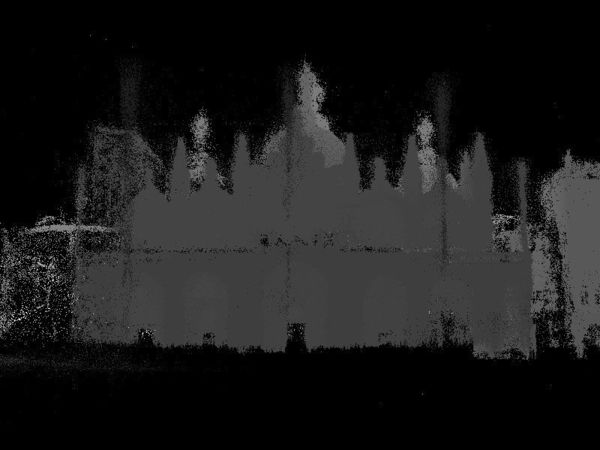}\\
        \includegraphics[width=\linewidth]{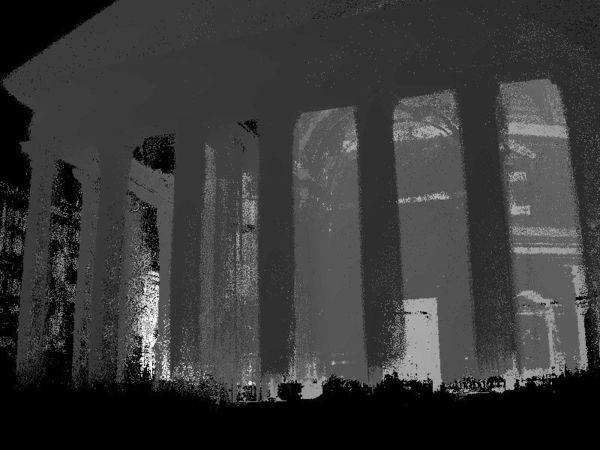}
        \caption{depth map}
    \end{subfigure}\,%
    \begin{subfigure}[b]{0.32\linewidth}
        \centering
        \includegraphics[width=\linewidth]{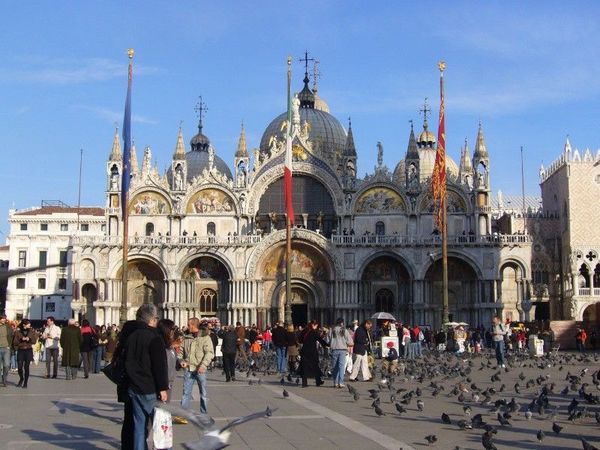}\\
        \includegraphics[width=\linewidth]{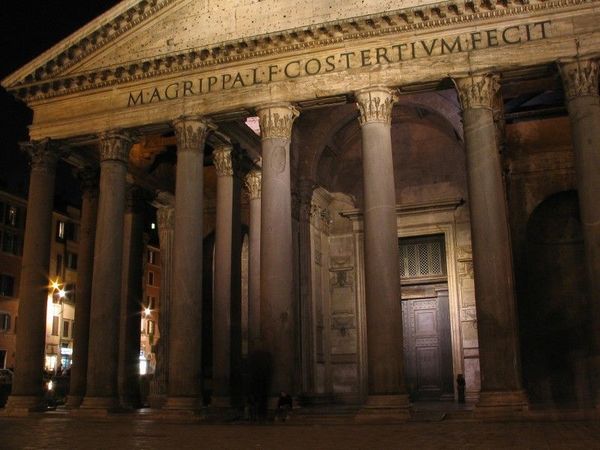}
        \caption{original image}
    \end{subfigure}
    \caption{Sample frames of the aligned dataset.
    Even though interior structures can be seen through the walls in the point cloud rendering (bottom), neural rerendering is able to reason about occlusion among the points and thereby avoid rerendering artifacts. \footnotesize{Image credits: James Manners, Patrick Denker (Creative Commons).}}
    \label{fig:aligned_dataset}
\end{figure}

\subsection{Baseline architecture}
We use a faithful Tensorflow implementation of the encoder-decoder network and appearance encoder in~\cite{lee2018diverse} using their PyTorch released code as a guideline.
We adapt their training pipeline to the single-domain supervised setup as described in Section 3.2 in our paper.

\subsection{Aligned datasets}
Figure~\ref{fig:aligned_dataset} shows sample frames from aligned datasets we generate as described in Section~\ref{sec:neural_rerendering_framework}.

\subsection{Latent space visualization}
\begin{figure*}[ht!]
    \centering
    \captionsetup[subfigure]{aboveskip=1pt,belowskip=3pt}
    \begin{subfigure}[b]{0.65\linewidth}
        \centering
        \includegraphics[width=\linewidth]{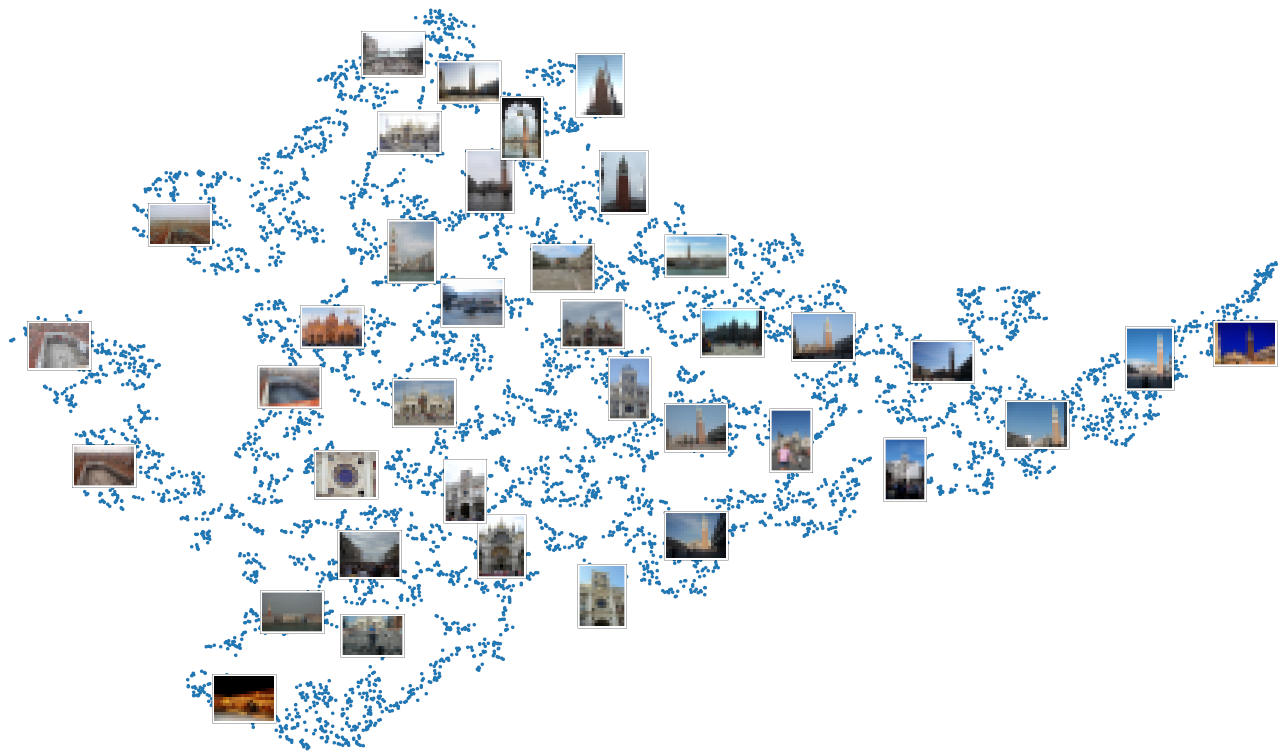}\\
        \caption{Our staged training: After appearance pretraining.}
        \label{fig:latent_vis_pretrain}
    \end{subfigure}
    \begin{subfigure}[b]{0.65\linewidth}
        \centering
        \includegraphics[width=\linewidth]{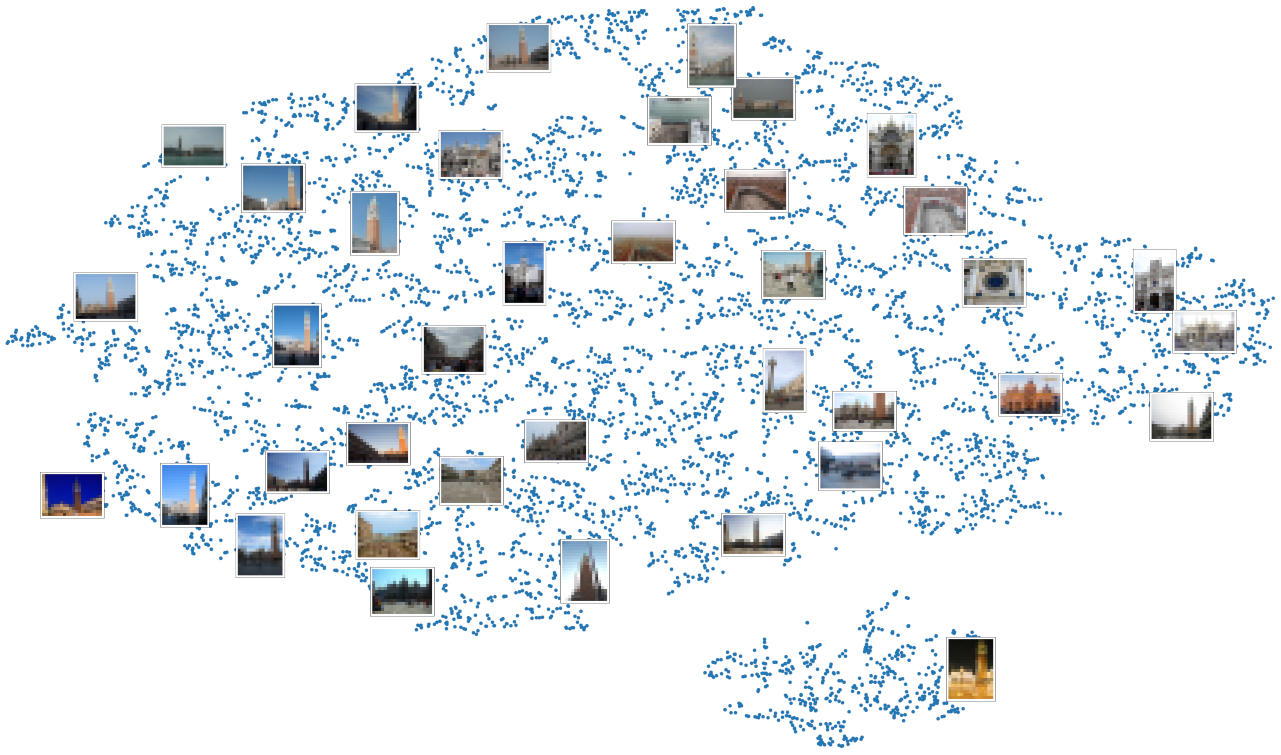}\\
        \caption{Our staged training: After finetuning.}
    \end{subfigure}
    \begin{subfigure}[b]{0.65\linewidth}
        \centering
        \includegraphics[width=\linewidth]{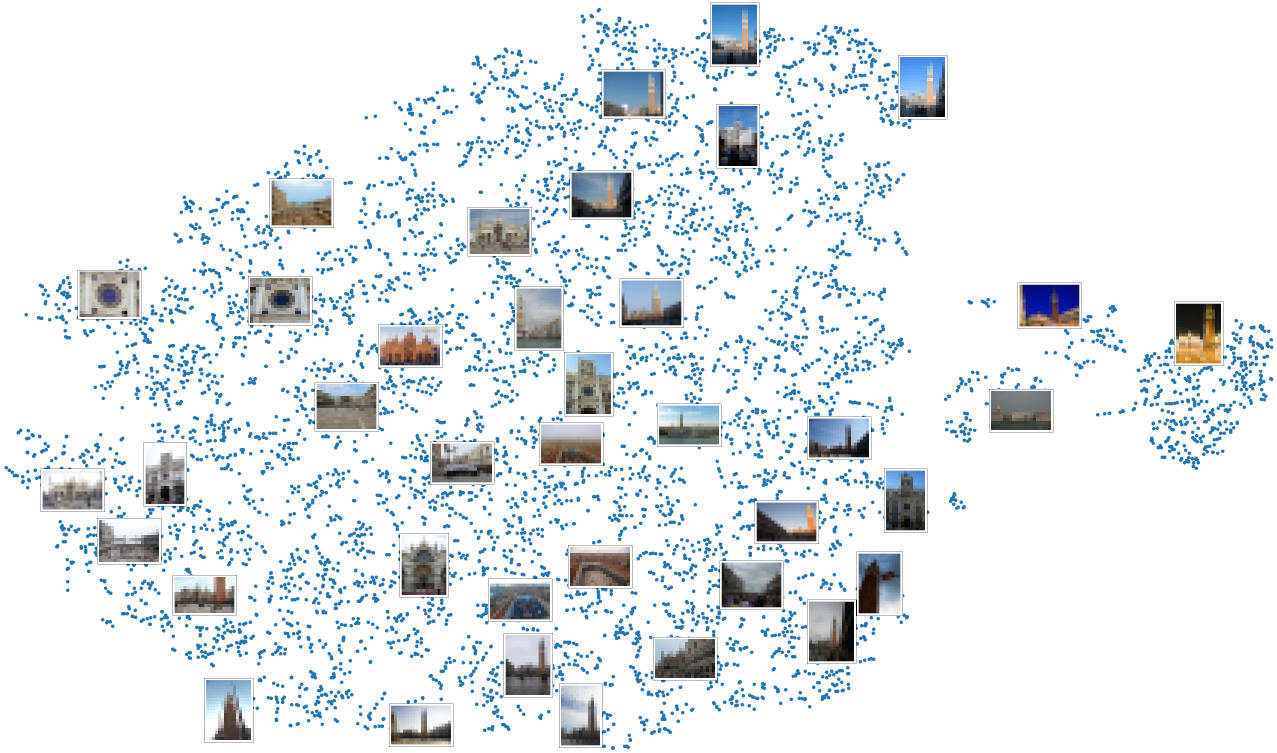}\\
        \caption{BicycleGAN baseline.}
    \end{subfigure}

    \caption{t-SNE plots for the latent appearance space learned by the appearance encoder (a) after appearance pretraining in our staged training, (b) after finetuning in our staged training, and (c) using the BicycleGAN baseline.}
    \label{fig:latent_vis}
\end{figure*}

Figure~\ref{fig:latent_vis} visualizes the latent space learned by the appearance encoder, $E^a$, after appearance pretraining and finetuning in our staged training, as well as training $E^a$ with the BicycleGAN baseline. 
The embedding learned during the appearance pretraining stage shows meaningful clusters, but has lower quality than the one learned after finetuning, which is comparable to the one of the BicycleGAN baseline.

\clearpage
{\small
\bibliographystyle{ieee}
\bibliography{references}
}

\end{document}